\documentclass[]{article}
\AtBeginDocument{%
  }

\usepackage{mathtools}
\usepackage[]{todonotes}
\usepackage{tabularx}
\usepackage[]{pdfcomment}
\usepackage[]{tikz}
\usepackage{orcidlink}
\usepackage{amsmath}
\usepackage{amssymb}
\newcommand{\eps}{\varepsilon}
\newcommand{\R}{\mathbb{R}}
\newcommand{\abs}[1]{\left|#1\right|}
\newcommand{\norm}[1]{\left\|#1\right\|}
\newcommand{\set}[1]{\left\{#1\right\}}
\renewcommand{\vec}{\mathbf}
\newcommand{\supp}{\mathop{\text{supp}}}

\newtheorem{example}{Example}

\newenvironment{acks}{\paragraph{Acknowledgement}}{}

\title{Statistically Testing Training Data for Unwanted Error Patterns using Rule-Oriented Regression}
\author{Stefan Rass \orcidlink{0000-0003-2821-2489}
	\thanks{Johannes Kepler University, LIT Secure and Correct Systems Lab, Linz, Austria, \\\href{mailto:stefan.rass@jku.at}{\texttt{stefan.rass@jku.at}}}
\and
Martin Dallinger \orcidlink{0009-0002-5293-2937}\thanks{
		Johannes Kepler University, Linz, Austria, email: \href{mailto:martin.dallinger@outlook.com}{\texttt{martin.dallinger@outlook.com}}}}

\begin{document}

\maketitle


\begin{abstract}
Artificial intelligence models trained from data can only be as good as the underlying data is. Biases in training data propagating through to the output of a machine learning model are a well-documented and well-understood phenomenon, but the machinery to prevent these undesired effects is much less developed. Efforts to ensure data is clean during collection, such as using bias-aware sampling, are most effective when the entity controlling data collection also trains the AI. In cases where the data is already available, how do we find out if the data was already manipulated, i.e., ``poisoned'', so that an undesired behavior would be trained into a machine learning model? This is a challenge fundamentally different to (just) improving approximation accuracy or efficiency, and we provide a method to test training data for flaws, to establish a trustworthy ground-truth for a subsequent training of machine learning models (of any kind). Unlike the well-studied problem of approximating data using fuzzy rules that are generated from the data, our method hinges on a prior definition of rules to happen before seeing the data to be tested. Therefore, the proposed method can also discover hidden error patterns, which may also have substantial influence. Our approach extends the abilities of conventional statistical testing by letting the ``test-condition'' be any Boolean condition to describe a pattern in the data, whose presence we wish to determine. The method puts fuzzy inference into a regression model, to get the best of the two: explainability from fuzzy logic with statistical properties and diagnostics from the regression, and finally also being applicable to ``small data'', hence not requiring large datasets as deep learning methods do. We provide an open source implementation for demonstration and experiments.

	
\end{abstract}

\section{Introduction}

In many cases the success of artificial intelligence depends on large amounts of training data to be available, but in addition to obtaining the desired quantity, the quality of the training samples is a separate question and challenge. The reproduction of undesired biases (such as discriminatory, racist, etc.) is a documented phenomenon \cite{hofmann_ai_2024,blodgett_llms_2024}, and considerable research has delved into ``de-basing'' training data. 

This work explores a method of obtaining an AI model that shares the explainability of rule-based AI, while remaining eligible to statistical diagnostics and quantitative accuracy measurement. The method was inspired by the challenge of automating data quality assessments with help of artificial intelligence. This work presents a re-implementation of the approach within an enhanced data pipeline, aiming to ensure trustworthy training data that minimizes undesired biases. By automating data quality assessments, this method not only enhances the reliability and trustworthiness of the data but also contributes to its reusability and interoperability, thereby aligning with the Findability, Accessibility, Interoperability, and Reuse of digital assets (FAIR) principles. To explain the intuition, it pays off to briefly review the past work and findings, whose generalization to other contexts is the goal of the present work.

\subsection*{Introduction by Example: Automating Data Quality Assessment}
In a past work \cite{cichy_fuzzy-approximation_2019,cichy_fuzzy_2020}, an application of regression analysis to automate the assessment of data quality was discussed, which was (up to this point) a matter of manual labeling. Experts were given tables with records, whose quality was to be assessed in terms of several indicators, summarized in Table \ref{tbl:quality-indicators}. The list given in this table is far from comprehensive as can be seen in \cite{cichy_overview_2019}, yet sufficient for the illustrative purpose of this example. Moving towards partial automation, it is not difficult to evaluate all these formulas in a spreadsheet, allowing for a subsequent subjective quality assessment, as previously done by experts. Handing this (still tedious) task over to an AI model is nothing but a natural next step, but comes with the requirement of explaining each assessment in case. Depending on the business sector, explainability can come as a stringent requirement. This work aims to address these requirements, balancing automation with interpretability to meet both the technical and regulatory needs of data-driven decision-making.

\begin{table}[b!] 
	\centering
	\caption{Subset of the variables relevant for data quality assessment} 
	\includegraphics[width=\textwidth]{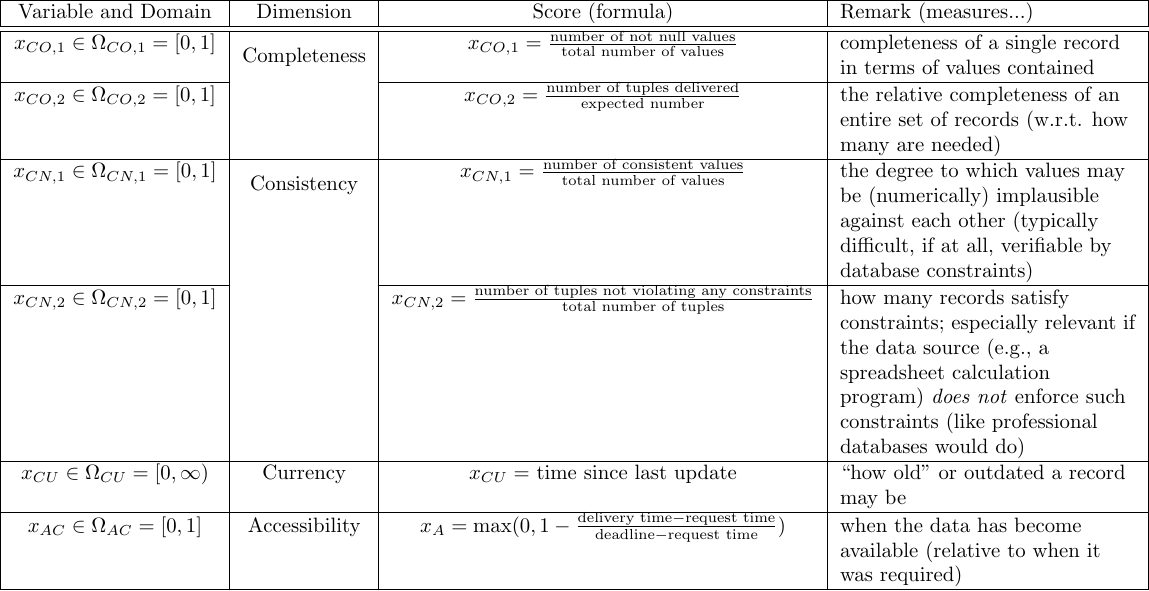}
	\label{tbl:quality-indicators}
\end{table} 

As a very simple explainable model, regression analysis was tested, but simply adding up the variables in a linear combination towards a 
\begin{align}
	\begin{array}{r}
		\text{data quality (DQ)}\\
		\text{prediction}
	\end{array} &= \beta_0 + \beta_1 \cdot x_{CO,1}+\beta_2 \cdot x_{CO,2}+\beta_3 \cdot x_{CN,1}+\beta_4 \cdot x_{CN,2}\nonumber\\
	&\qquad + \beta_5\cdot x_{CU}+\beta_6\cdot x_{AC}\label{eqn:simple-regression-model}
\end{align}

and fitting the $\beta$-coefficients with the standard techniques would work, but leaves much to be desired about explainability. Some of the variables, like completeness and consistency, share the same units (e.g., dimensionless fractions), but others, such as $x_{CU}$ (used to measure the timeliness of records), carry units. Simply assuming the respective $\beta$-coefficients with the corresponding ``inverse'' unit to make the addition well-defined is only bypassing but not solving the issue, which is why fuzzy rules have been proposed.

The improvement proposed by \cite{cichy_fuzzy-approximation_2019} was to replace the ``plain'' variables in \eqref{eqn:simple-regression-model} by a set of continuous (basis) functions $f_i:\Omega_{CO,1}\times\Omega_{CO,2}\times\cdots\times\Omega_{AC}\to[0,1]$ (as data quality is measured on the scale from 0 to 1), which in turn can all depend on all variables of interest, and embody an arbitrarily complex logic among the variables to determine the actual data quality. In other words, the heuristic according to which a human expert may set the subjective data quality assessment can for instance take the following form:
\begin{equation}\label{eqn:rule-based-assessment}
		\left.\begin{array}{l}
		\text{\textbf{if}~} x_{CO,1}\geq 0.7\text{~\textbf{then}~the data quality is \emph{high}}\\
		\text{\textbf{if}~} x_{CO,1} \in [0.3,0.7)\text{~\textbf{then} the data quality is \emph{medium}}\\
		\text{\textbf{if}~} x_{CO,1} <0.3\text{~\textbf{then} the data quality is \emph{low}}
	\end{array}\right\}
\end{equation}
Rules of the form \eqref{eqn:rule-based-assessment} are typically what a decision-tree analysis would produce, with the main algorithmic labor going into finding the optimal threshold values (like 0.7 and 0.3 above). The actual logic then takes the form of a nested reasoning, taking decisions in a certain order of the variables, and iteratively refining the final decision. Randomizing this process to compile the final result from a multitude of different trees leads to \emph{random forests}, which are known for often very good accuracy, however, at the price of losing the explainability to some extent (since the rules become ambiguous if they appear in several trees in slightly different form).

Fuzzy-rule based regression further modifies the approach by combining \eqref{eqn:simple-regression-model} with pre-specified rules like \eqref{eqn:rule-based-assessment}, while each set of rules is integrated into its own continuous function $f_i$ that takes (some, up to all) variables of interest to a continuous basis function inside a regression model. In addition, fuzzy logic allows to use natural language terms to replace otherwise crisp thresholds or intervals, so we may rewrite \eqref{eqn:rule-based-assessment} again in more natural language terms and compile it into a continuous function by standard methods (see, e.g., \cite{ross_fuzzy_2002} for a detailed account):
\begin{equation}\label{eqn:fuzzy-rules-example}
	\left.\begin{array}{l}
		\text{\textbf{if}~} x_{CO,1} \text{is \emph{high},~\textbf{then}~ DQ is \emph{high}}\\
		\text{\textbf{if}~} x_{CO,1} \text{~is \emph{med.}, \textbf{then}  DQ is \emph{medium}}\\
		\text{\textbf{if}~} x_{CO,1} \text{~is \emph{low}, \textbf{then}  DQ is \emph{low}}
	\end{array}\right\}\xRightarrow[\text{known techniques}]{\text{conversion by}} f_1: \Omega_{CO,1}\to[0,1]
\end{equation}
A rule-set like exemplified in \eqref{eqn:fuzzy-rules-example} utilizes only one among six variables mentioned in Figure \ref{tbl:quality-indicators}, yet it is straightforward to imagine any set of more expressions to be included in the construction of the function $f_1$ or even to define the function $f_1$ over the entire domain of the vector $\mathbf x$, depending on the requirements of the application. Each such function is a small-scale expert system carrying its input arguments $\vec x$ through the usual process of $\vec x\to$ fuzzification $\to$ inference $\to$ defuzzification $\to f(\vec x)$. Using these as basis functions, we can set up a conventional regression model likewise as
\begin{equation}\label{eqn:fuzzy-rule-based-regression}
\text{data quality} = \beta_0 + \beta_1\cdot f_1(\vec x)+ \beta_2\cdot f_2(\vec x)+\ldots+ \beta_m\cdot f_m(\vec x)+\eps,	
\end{equation}
in which each $f_i$ embodies another set of logical heuristics, and $\vec x$ is the entire vector of variables of interest, and $\eps$ is the error term (with accompanying assumptions on its distribution, correlation and moments).  The final (fitted and filtered) model can, roughly, be considered as a form of Takagi-Sugeno-Kang reasoning, if we take the entirety of if-then rules disjunctively connected to output a function of the input variables, taking the general form as in \eqref{eqn:fuzzy-rule-based-regression}. However, following the advice of \cite{mendel_2021_critical}, we may not consider the result as a crisp logical rule-set to explain the data, but rather as flexible basis functions designed for interpretability, and whose statistical significance can reveal patterns in the data. 

Observe that the number $m$ of basis functions here depends on how many sets of heuristic rules we may define a priori, and their definition in turn requires a pre-defined ``vocabulary'' of linguistic variables to give terms like ``low'', ``medium'' or ``high'' a meaning in their individual contexts (noting that ``low'' is used twice, but referring to the variable $x_{CO,1}$ and the data quality in \eqref{eqn:fuzzy-rules-example}, both of which can have quite different meanings). As a technical convenience, observe that \eqref{eqn:fuzzy-rule-based-regression} is still a \emph{linear} model, although the basis functions themselves can be quite nonlinear. Setting aside the technical details of converting textual rule specifications into fuzzy membership functions (which is what's meant by ``known techniques'' in \eqref{eqn:rule-based-assessment}), as well as the process of fitting the $\beta$-coefficients in \eqref{eqn:fuzzy-rule-based-regression} by least-squares or other methods, let us instead highlight a few findings that we can read off the model directly, once obtained:
\begin{itemize}
	\item The \emph{magnitude} $\abs{\beta_i}$ in \eqref{eqn:fuzzy-rule-based-regression} is interpretable as an importance indicator, since the more weight a rule receives in the fitting, the more variation in the data could be explained by it.
	\item The \emph{sign} of each $\beta_i$ indicates a positive or negative correlation between the rule, if it fires, and the predicted variable. To interpret this case correctly, bear in mind that the actual value of a rule $f_i(\vec x)$ that embodies a (set of) ``\textbf{if} \ldots \textbf{then} $x$ is \emph{some ordinal value}'' is an approximation of how the output (predicted) variable is affected if the conditions of the antecedent are met. If, under these conditions, the rule fires, then a, for example higher, value of $x$ would negatively correlate to the value that it should be according to the training data. This would suggest the possibility of considering a version of the rule(s) behind $f_i$ with a consequent that could be revised to the negation of the current version (particular care is in order for attributes whose value set is more than binary, since a negative weight for a rule that says ``\ldots \textbf{then} $x$ is \emph{low}'' would, after a revision, need a consequent like ``\ldots \textbf{then} $x$ is not \emph{low}'', leaving several possible values to be ``not \emph{low}'', which complicates the rule and its interpretation).
	
	In any case, a negative coefficient $\beta_i$ for rule $f_i$ may indicate either a mis-specification of the terms (used to construct $f_i$) or a problem in the labeling that the regression model approximates. Specifically, it suggests that the rule would, in cases where it ``fires''\footnote{Technically, a fuzzy rule can, by construction, always come to execution, but its ``strength'' in terms of membership can become arbitrarily low, up to zero, which corresponds to the rule ``not firing''.}, predict the data quality to be increasing under the conditions of the rule, while the data shows the exact opposite, as indicated by the sign of the $\beta$-coefficient.
\end{itemize}
The actual model, quality of fit and value of coefficients therein, will strongly depend on how the linguistic variables are defined, over which the rules are formulated. The many possibilities to choose a triangular norm (and co-norm) to model logical connectives in Boolean expressions (at least for the implications into which if-then rules like the above are converted into), will also affect the quality of the fitted model. We leave the exploration of these factors as an aisle of future study, and instead consider the ``inner logical consistency'' of the rules themselves, using the standard $\min,\max$ and $x\mapsto 1-x$ to represent the Boolean conjunction, disjunction and negation operations.

It is nonetheless important to remark that prior research \cite{mendel_2021_critical} found it not valid to explain the output of Mamdani or Takagi-Sugeno-Kang rule-based fuzzy systems using if-then rules. To avoid this mistake in our work, we are explicitly careful to \emph{not} treat the rules as logical implications besides their role as basis functions; rather, we use the correlations between variables (in certain combinations formulated as rule antecedents) with the given response variable. Nonetheless, the logical appearance of the test patterns is indeed admissible as an approximation of a logical implication, as long as we are careful not to do (crisp-logical) reasoning with the fitted rules (which we avoid). We come back to this issue during Section \ref{sec:evaluation} where we give examples, and in the discussion (Section \ref{sec:discussion}).

Suppose that the data is well explained by a fuzzy model (and the past work in \cite{cichy_fuzzy-approximation_2019} found the fuzzy-rule based regression to outperform a standard regression), but the rules are, when treated as in two-valued logic, found as inconsistent with each other.  This is exactly the kind of pointer towards errors in the training data that we are looking for (and may not necessarily be obtained by data-reliant rule generation methods) . More concretely, suppose that two rules are found significant, which differ only in the consequent like in the following example:
\begin{itemize}
	\item $f_1:$ \textbf{if} accessibility is \underline{\emph{medium}} \textbf{then} the data quality is \underline{\emph{high}}
	\item $f_2:$ \textbf{if} accessibility is \underline{\emph{medium}} \textbf{then} the data quality is \underline{\emph{low}}
\end{itemize}


While in this example, rules are obviously mutually contradictive, such a discovery in the training data is nontrivial, since the respective rules are typically not explicitly expressed, known or constructed in a standard machine learning data pipeline. However, if inconsistent rules like the above are discovered as statistically significant in a regression model, then this allows some hypotheses about the training data in first place, as for example (but not limited to):

\begin{itemize}
	\item Several people could have been labeling the data with perhaps opposite understandings of what ``accessibility'' means or what this implies about the data quality.
	\item The same person could have made a humble mistake by looking into one record, but accidentally labeling another record instead (for example, if the labeling was done manually in a large spreadsheet, and the cursor jumped into another row without the person noticing).
	\item Other reasons, like fatigue, stress, or similar factors, may contribute to inconsistent labeling behavior.
\end{itemize}

The remainder of this work is dedicated to a discussion and position about combining fuzzy logic with statistical regression and logical reasoning to discover unwanted patterns in data used to train artificial intelligence. We leave the introductory example here, and turn to the general data pipeline that we propose as a method to uncover errors in training data, inspired by the above discussion.

\section{Related Work}
The challenge of ensuring data quality and mitigating biases in AI training datasets has been the focus of extensive research. Various approaches (from at first sight rather unrelated areas) have been explored to identify unwanted patterns, so we divide our review in sections that cluster different approaches.

\subsection{Criticality of Bias Detection in AI and Debiasing}
Identifying biases within AI systems is critical for ethical machine learning, primarily as the resulting decisions can be impactful. Barocas et al. \cite{barocas_2023_fairness} discuss inherent technical biases during the data preprocessing stage, highlighting the need for fairness and transparency in AI practices, as does a significant body of related research \cite{Ekstrand0B022, RajE22, wang_factors_2020, YaoH17, AshokanH21, Baeza_Yates18}. They argue that fairness is hard to patch into systems post-implementation but should be a central concern for the entire development process, starting with the data. Our work falls into the category of pre-processing methods for de-biasing, which currently works by bias-aware sampling \cite{Kirnap0BECY21, melchiorre2021investigating}, or bias measuring \cite{RekabsazKS21, Rekabsaz0HH21}, for which we describe a new method here.  For the subsequent task of repairing the dataset \cite{Heidari_2019, Mahdavi}, we contribute a  heuristic on how to identify records that contribute to the bias.

Various de-biasing techniques have been proposed in the literature \cite{gonzalez-sendino_mitigating_2024,zhang_mitigating_2018}, including re-weighting, resampling, adversarial de-biasing, and others. Common to these methods is the assumption of prior knowledge about under-represented elements in the sample (such as possibly underrepresented groups of individuals or similar), whose representation in the training data can be corrected in accordance to this knowledge. Our work aims at the prior point in time where we seek to discover such biases without knowing them already. Similar methods \cite{ferrara_fairness-aware_2024} aiming at proper pre-processing or sampling subsets of the training data to remove biases (fairness in data selection and preparation, model design, evaluation and testing) likewise depend on knowledge about the bias' existence, whose discovery is the main challenge that we aim at.

\subsection{Regression-Based Explanations}
Regression analysis offers both predictive power and interpretability. While traditional regression models rely on linear relationships, well-known techniques such as \emph{classification and regression trees} (CART) \cite{death_2000_CARTAP} also provide flexibility and explainability in uncovering complex patterns, but have binary (non-fuzzy) if-then constraints and are usually simplified with pruning to get a reasonably complex fit. Other than this work, hybrids combining fuzzy with other AI and machine learning techniques have been proposed \cite{chimatapu_explainable_2018}, including fuzzy-neural systems, fuzzy-belief networks, and many others; the application within regression \cite{cichy_fuzzy-approximation_2019,cichy_fuzzy_2020}, however, has widely unexplored potential.

Another prominent method for explaining data with many possible causes is the \emph{least absolute shrinkage and selection operator} (LASSO), which enhances regression by simultaneously performing variable selection and regularization. This technique imposes an $\ell_1$-norm penalty, shrinking coefficients of less significant variables toward zero, thus yielding sparse, interpretable models even in high-dimensional datasets \cite{tibshirani_1996_RegressionSA}. Its ability to identify relevant predictors, even with significance tests as shown in \cite{geer_asymptotically_2014}, while reducing overfitting has made it a cornerstone in modern statistical modeling.

\subsection{Fuzzy Logic and Rule-Based Systems}
Fuzzy logic systems have been employed to enhance interpretability in AI models, achieving a balance between crisp rules and statistical models. Ross \cite{ross_1994_FuzzyLW} provides a comprehensive overview of fuzzy logic principles, integrating fuzzy systems into AI applications to manage uncertainties, and Fumanal-Idocin et al. \cite{fumanal-idocin_ex-fuzzy_2024} provide a fuzzy library for symbolic explainable AI. 


\subsection{Rule Mining for Data Pattern Discovery}\label{sec:rule-mining}
Mining logical rules from data has proven effective for identifying latent structures and biases. Techniques such as association rule learning and decision tree induction have been adapted for extracting compact, human-readable rules \cite{aggarwal_2015_DataMT}. Our approach uses some of these methods as a heuristic for interestingness to pre-filter automatically generated rules.

It is important to delineate our work from related methods like the Wang-Mendel approach \cite{wang_generating_1992}, which also attempts to explain the data in terms of fuzzy rules, fuzzy regression \cite{chukhrova_fuzzy_2019} or general trustworthy XAI methods \cite{chamola_review_2023}.  While the latter clearly point out regression as one well-explainable method, using fuzzy logic as basis functions for regression was put forth in \cite{cichy_fuzzy-approximation_2019}, but to the best of our knowledge not yet deeper studied for XAI. Fuzzy regression techniques, in turn, mostly generalize different forms of regression to using fuzzy variables or to capture uncertainty in the estimation with fuzzy techniques. Our application of fuzzy logic differs mostly in the purpose of the ``fuzzyness'' as such; the benefit of fuzzy logic in our case is created from their provisioning of a continuous function to represent a logical if-then rule. In the form of a function, an if-then rule can serve as a basis function in a regression-based data approximation. To the same end, the celebrated Wang-Mendel method is closely related. In its basic form (see \cite{wang_generating_1992}, on which state-of-the-art versions are based \cite{lapa_increasing_2024,huhn_furia_2009,bollaert_frri_2025}), this method creates one rule per data record, and iteratively reduces the so-created initial set of rules until the smallest set is left that still explains the data sufficiently well. The important difference to our approach is that the Wang-Mendel algorithm would find patterns only if they are ``directly expressed'' by a particular record; in contrast, we look for patterns that are implicitly encoded by the presence of multiple records, among none of which would indicate an undesired pattern by itself (in isolation).

Independently and more importantly, our problem is more similar to statistical hypothesis testing, where the proper procedure prescribes to formulate the hypothesis \emph{before} the data is collected, and to design the experiment \emph{based} on the hypothesis to test. Reversing this sequence entails various risks (like working with implicit assumptions about the data when defining the rules, confirmation biases, inadvertent data dredging, reduced credibility in terms of predictive power, etc.). Our setting is such that the data already exists, but formulating the pattern to look for based on the data, like the Wang-Mendel method would do, bears similar risks as formulating test hypotheses a posteriori. 

\section{The Data Pipeline to Uncover Undesired Data Patterns}
\begin{figure}[htbp]
    \centering
    \begin{tikzpicture}
        \node at (0,0) {\includegraphics[width=\textwidth]{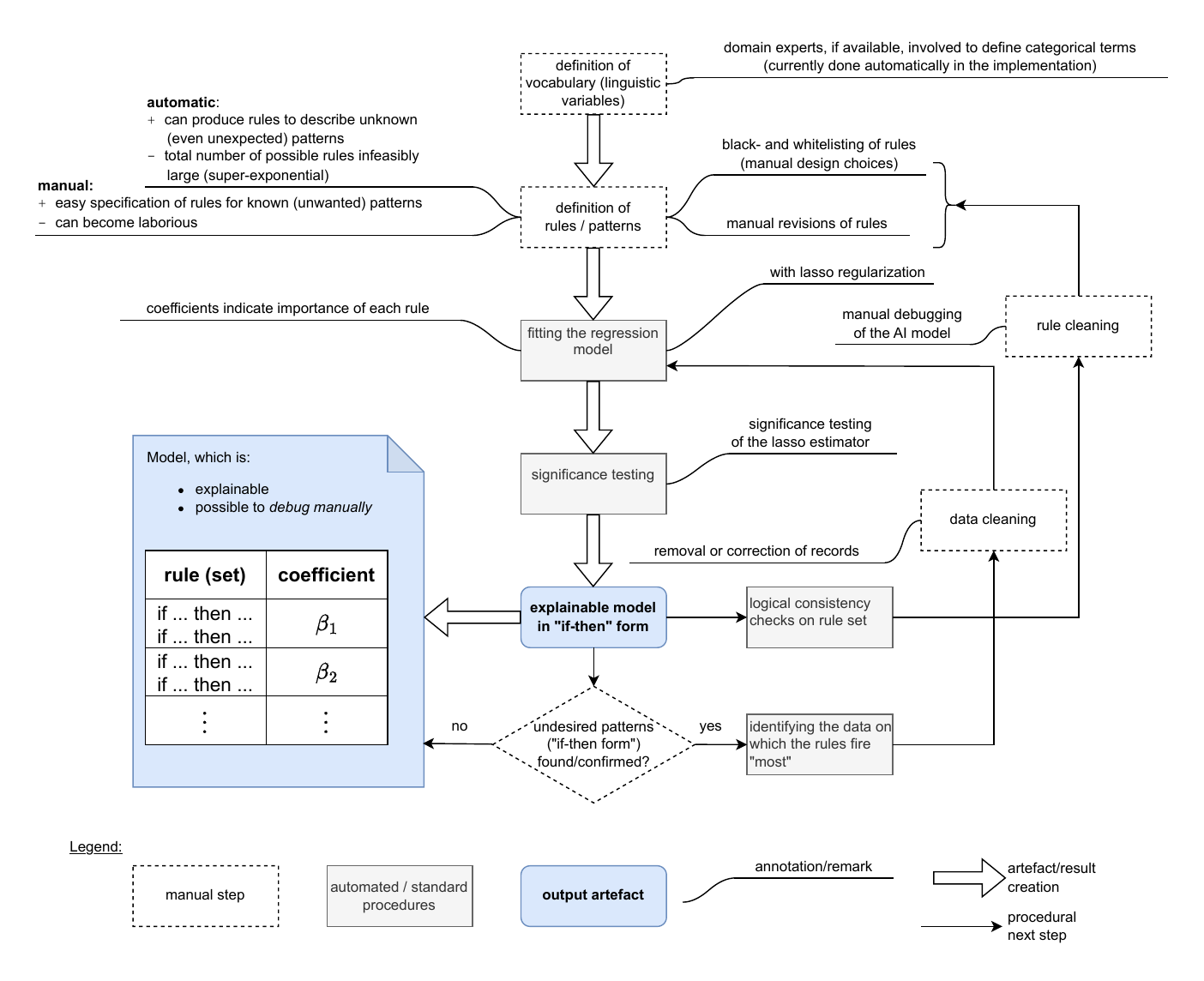}};
        \node at (0.5, 1.2) {\scalebox{0.5}{\cite{geer_asymptotically_2014}}};
        \node at (0.5, -0.1) {\scalebox{0.5}{\cite{geer_asymptotically_2014}}};
        \node at (2.2, -1.5) {\scalebox{0.4}{\textsf{(Python $\to$ footnote \ref{ftn:logical-checkup-script})}}};
        \node at (2.4, -2.75) {\scalebox{0.4}{\textsf{(see Sect. \ref{sec:rule-tracing})}}};
    \end{tikzpicture}
    \caption{Pipeline to explain data from logical patterns}
    \label{fig:workflow-gesamt}
\end{figure}

Figure \ref{fig:workflow-gesamt} shows an overview of the workflow, whose details we discuss below. Steps that follow standard procedures of logic or statistics are hereafter not expanded further, unless needed to emphasize particularities or changes that are important in a sub-step.

\newcommand*{\xMin}{-6}%
\newcommand*{\xMax}{6}%
\newcommand*{\yMin}{-6}%
\newcommand*{\yMax}{6}%


The basic idea is to describe the data using a set of heuristic rules that we assume can effectively capture its underlying structure. To this end, we need to determine:
\begin{itemize}
	\item The linguistic terms (Section \ref{sec:automated-rule-generation}) in which we formulate the rules. This step is technically needed as the first, but the terms may be identified along the determination of the rules (hence, the two steps are interwoven).
	\item The rules themselves (Section \ref{sec:automated-rule-generation}): manually, automatically or semi-automatically, but in any case, based on the hidden patterns that we are looking for. 
	\item The statistical significance of each rule (Section \ref{sec:sig-testing}), i.e., its contribution to the explanatory power of the regression model. This is up to suitable statistical tests \cite{geer_asymptotically_2014}.
	\item The logical consistency of the rules (Section \ref{sec:discussion}), to identify contradictions or other inconsistencies, enabling subsequent adjustments to the training data to address biases or correct inaccurate labels. 
\end{itemize}

Our exposition in the following will not go into technical details on how fuzzy sets are to be defined, or how the expert systems are technically created, since this is (also in our implementation) left up to standardized procedures. Our method is agnostic to how the linguistic variables, fuzzy inference rules and other aspects are designed, and requires only a way of turning logical statements into continuous functions to use inside a standard regression (whether this is done by fuzzy logic or by other means is secondary here). Although the results of the method will be highly dependent on the quality of the foregoing fuzzy modeling, our aim is here \emph{not} to provide a newer and better fuzzy expert system design for a certain bias modeling in data, but rather the proposal of a new method to uncover biases, \emph{assuming} that their general expected ``appearance'' in the data can be modeled well. We will thus uncover technicalities of the fuzzy building blocks only to the extent of needing them to highlight certain particularities about the showcased models, or how the results are to be interpreted.

The \emph{data} that we start with is considered a set of records, each of which is a tuple that binds together data of different types. To simplify matters here, we shall explicitly discuss only the case of numeric and non-numeric (e.g., categorical or textual) information. In case of data about people, this could be age (numeric), income (numeric), but also gender (textual), heritage (textual), and others. To compactly refer to the data hereafter, we let it come as a data matrix $\vec D$, with $n$ rows, called \emph{records}, and any number of columns, that we will call \emph{fields}. Among the columns, we single out one that carries the information on which some AI model shall be trained on, and we divide a record $\vec d\in \vec D$ into a vector $\vec x=(x_1,\ldots,x_k)$ of independent variables, and a response value $y_i$, so that $\vec d_i=(\vec x_i,y_i)$.

To keep track of the variables which will be introduced incrementally coming in the following sections, let us summarize them in Table \ref{tbl:list-of-symbols}, to make the exposition easier to follow.

\begin{table}[h!] 
\centering
\caption{List of symbols}\label{tbl:list-of-symbols}
\begin{tabular}{|c|>{\raggedright}p{8cm}|}
	\hline 
	Variable & Meaning\tabularnewline
	\hline 
	\hline 
	$\vec{D}$ & data, set of records\tabularnewline
	\hline 
	$n$ & number of records in $\vec{D}$\tabularnewline
	\hline 
	$\vec{d}_{i}=(\vec{x_{i}},y_{i})$ & $i$-th data record, with independent variables $\vec{x_{i}}=(x_{1},\ldots,x_{k})$
	and single response variable (label) $y_{i}$\tabularnewline
	\hline 
	$k$ & number of independent variables \tabularnewline
	\hline 
	$f_{j}$ & a single basis function that embodies one or several if-then rules
	that (together) are a logic description of a pattern that we seek
	to uncover\tabularnewline
	\hline 
	$m$ & number of basis functions, including the (constant) intercept \tabularnewline
	\hline  
\end{tabular}
\end{table}

\subsection{Defining the Vocabulary and Rules}\label{sec:automated-rule-generation}
Given the data table $\vec D$, the first step is defining linguistic terms to describe the allowed values of variables in $\vec D$. This can, already, be made in terms that allow a direct definition of patterns that we are interested in finding. For example, if the data at hand is suspected to carry discriminating, unethical or other undesirable patters like ``women of a certain age range should not be employed, because they may become pregnant and temporarily drop out of the job'', our vocabulary can already be defined in the respective terms.

\begin{example}[Family planning discrimination in a hiring process]\label{exa:family-discrimination}
	If we have information about the age $a$, we can divide the range $[0,a_{\max}]$ into three parts, being $[0,20]$ for young women, $[15,45]$ for medium age, and $[40,a_{\max}]$ for older women. Observe that this division is already questionable and could open ethical discussions, but remember that we are here explicitly looking for codifications of such unwanted thinking in whoever labeled the dataset $\vec D$ in first place. Hence, although ethically questionable, we \emph{should} anticipate such considerations to define a rule that can later capture the pattern accordingly, in addition to the patterns that we need to describe the data as such (e.g., based on the hypothesis about how the data arises). Building on this division of the age range, we can (again unethically, but tailored to the purpose), define linguistic variables like ``young'', ``capable of childbearing'' and ``old enough'' (to have finished family planning or no longer capable of doing so) to formulate a rule that could say
	\begin{equation}
		\text{\textbf{if} age is \emph{capable of childbearing} \textbf{then} employment is \emph{unlikely}}.
	\end{equation}
	Here, the new term ``unlikely'' comes in, which in turn requires a likewise (subjective) definition of what this means in practice (within $\vec D$).
	
	While this example is oversimplifying matters, real-life evidence of how anticipated fertility or risk of dropout due to care-taking obligations exists, and directly lends itself to finding terms like the above \cite{becker_discrimination_2019}, or other biases \cite{DBLP:journals/mansci/LambrechtT19,bonilla2006racism} (to mention only a few). In the evaluation Section \ref{sec:evaluation} we will construct such an artificially biased dataset, and demonstrate how the method discloses the bias, without receiving any prior or auxiliary information on the bias or its existence even. 
	
	
	\hfill $\blacksquare$
\end{example}

The overall goal of the first step is to define the model in terms that well describe the data and the logic to explain it. These include speculative rules to explain the data, explicit rules to describe unwanted patterns, and further (e.g., randomly generated) rules that may explain the data but have not yet been anticipated as such. 

Observe that we can, without loss of generality, confine ourselves to generating rules that only have AND operators and we neither need a negation nor a disjunction. The former is expressible by defining the respective complement sets, while the disjunction naturally comes in by having several rules available to explain the data.  Our choice to work with type-1 fuzzy sets here is arbitrary, and straightforwardly generalizable to type-2 or higher type fuzzy sets. Such generalizations naturally come in when the vocabulary definition is made collaboratively by several experts, each contributing the own (subjective) understanding of terms. In turn, patterns defined with included uncertainty about the linguistic variables may be more expressive, but at the same time require their own empirical data collection and respective construction. Precisely, when looking for a specific pattern that would be unethical or otherwise undesired, it would demand a designated empirical study and data collection to align the rules with domain expertise. This problem is put outside our scope here, since from the analytic viewpoint of our method, whether we use type-1 or type-2 methods is only a matter of accuracy in the data approximation, since the only requirement is a continuous function that represents a rule, no matter how constructed. Hence, for simplicity of the experimental evaluation, we may, again w.l.o.g., use type-1 fuzzy sets in this work.

Once the vocabulary and rules are defined, we can compile the resulting sets of rules into basis functions $f_1,\ldots,f_m: \R^k \to \R$, assuming that all categorical and continuous variables admit a representation over the reals. This process is based on a few assumptions:
\begin{itemize}
\item[A1:] Each variable $x_j$ in a record $\vec d_i$ is processed, based on only a finite number of categories into which the concrete (crisp) value can fall. For numeric variables, we hence need to (fuzzy-)partition the admissible range accordingly (as sketched in Example \ref{exa:family-discrimination}, or graphically displayed in Figure \ref{fig:example-linguistic-variable} for our custom ``biased salaries'' dataset \cite{dallinger2024}). 

Our choice of partitioning is for simplicity only, leaving a reasonable default to be replaced by a more informed definition of linguistic variables by domain experts or leveraging more advanced methods (e.g., \cite{comas_interval-valued_2025} that uses enhanced fuzzy-C means clustering), if available. However, we emphasize that the definition of membership functions should, for the sake of hypothesis design, be done independently (and before seeing) the data, for the reasons discussed in Section \ref{sec:rule-mining}.

Textual values are either directly representing a respective category (in which case a fuzzification can become trivial by assigning 100\% membership to the respective value), or be based on features of the respective text (to be computed by other means that we do not deepen here; as an example, matching a person's name against a list of (un)desired names, or assigning people based on their home address, country, or similar).

\begin{figure}
	\centering
	\includegraphics[scale=0.1]{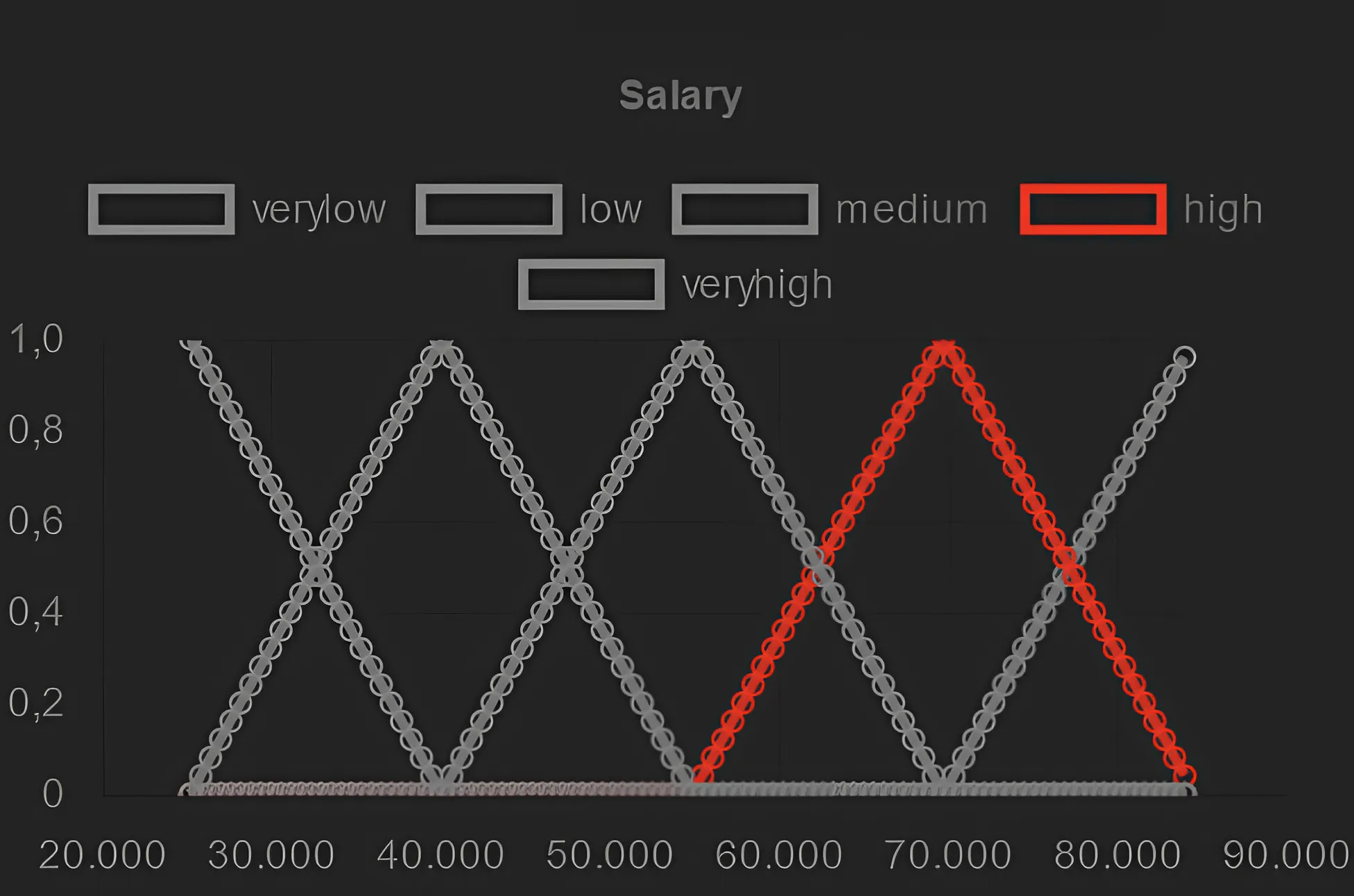}
	\caption{Example of fuzzy linguistic terms for the consequent variable ``Salary'' in our demo dataset \cite{dallinger2024}, with highlighted membership function for \ldots \textbf{then} Salary is \emph{high}}\label{fig:example-linguistic-variable}
\end{figure}

For continuous variables, we assume the values to be \emph{ordered}. Categorical variables (e.g., gender) are not ordered, and one-hot encoded by default. Variables that are to be taken as ordinal (i.e., finitely many but ordered values) must be encoded by numeric values before loading them into the program, hence the assumption about categorical values is made without loss of generality.

\item[A2:] The number of basis functions $m$ is much smaller than the number of samples $n$. Otherwise, if $m\geq n$, we could directly get to 100\% accuracy of the model by turning each record into a rule on its own, making the model overfit the data. A case with $m\geq n$ may, however, also naturally occur if we let the rules be generated automatically. This makes sense if we seek to discover some underlying logic about the labeling of the data, without providing any explicit hypotheses (rules) a priori. The case of generating rules automatically is discussed later. We remark that the condition $n\gg m$ is not a hard prior requirement, since the fitting process (Section \ref{sec:fitting}) includes a regularization step to reduce the set of rules without being affected by linear dependence of the rows or columns in the design matrix.

\end{itemize}

To define the terms for the writing of the rules to be searched for in the data, we simply:
\begin{enumerate}
	\item Iterate over all independent variables (columns) in the given dataset,
	\item and \textbf{for each} variable therein, 
	\begin{enumerate}
		\item \textbf{define} a \textbf{finite set of categories} into which values may fall. For continuous numeric ranges, these categories are ordered and may be (overlapping) intervals; for already discrete variables, their categories are just taken as the set of admissible values. For generally unstructured data, the categories will be defined based on features extractable from a text (if any), or otherwise be ignored for the further analysis.
		\item \textbf{define} a \textbf{membership function} for each category, treated as fuzzy sets into which values can fall. Our implementation (Section \ref{sec:linguistic-variable-def}) does this automatically using triangular membership functions.
	\end{enumerate}
\end{enumerate}
As for the rule generation, we may do this\ldots
\begin{description}
	\item[\ldots manually:] Here, we self-author rules according to patterns that we desire or that we hypothesize to explain the data well (hoping the respective rule to be among the significant ones), and patterns that are undesirable, as in Example \ref{exa:family-discrimination}. Once the model is fitted and tested for significance, it may be necessary to revisit this step to refine the model further.
	\item[\ldots automatically:] Like before, we may iterate over each variable, and create one if-then rule for each value it can take. For example, if the variable $x_1$ can take three values as \emph{low}, \emph{medium} or \emph{high}, and $y$ is the response variable (ranging over the same scale here, for simplicity, but not necessarily sharing the same meaning of the terms), we would create a block of $3\times 3=9$ rules:
	
	\begin{tabular}{lllll}
		\textbf{if} & $x_1=$ \emph{low}    & \textbf{then} & $y$ is \emph{low}    &  \\
		\textbf{if} & $x_1=$ \emph{medium} & \textbf{then} & $y$ is \emph{low}    &  \\
		\textbf{if} & $x_1=$ \emph{high}   & \textbf{then} & $y$ is \emph{low}    &  \\
		\textbf{if} & $x_1=$ \emph{low}    & \textbf{then} & $y$ is \emph{medium} &  \\
		$\vdots$    &  \\
		\textbf{if} & $x_k=$ \emph{high}   & \textbf{then} & $y$ is \emph{high}   &
	\end{tabular}
	
	All rules contain two variables, both with a range of $v$ values. Hence, we get a total of $O(v^2)$ many rules by this process, if the rules use only a single variable in their hypothesis. While this is still quadratic, it is about the practical limit of what can feasibly be constructed automatically, since even in case of only two values per variable, the total number of rules over $k$ variables equals the number of Boolean functions in $k$ variables would grow infeasibly large. 
	
	
	So, an automatic generation of rules will require systematic reductions towards a reasonable tradeoff, and modern methods based on the Wang-Mendel method \cite{lapa_increasing_2024,bollaert_frri_2025} would accomplish this efficiently and with good accuracy, but create the rules \emph{from} the data, and not a priori (cf. Section \ref{sec:rule-mining}). 
	For the latter, orthogonal arrays \cite{hedayat_orthogonal_1999} appear as a natural mechanism to produce rules systematically with a reasonable cover of combinations of variables (more classical methods like Wang-Mendel algorithm are here explicitly not adopted, since the production of rules should happen independently of the data). However, it may happen (as was experimentally observed for the data in Section \ref{sec:biased-salaries-example}) that rules that could explain an undesired bias were, by coincidence, not among the candidates. The idea of letting users of the method self-define their vocabulary to aid interpretability has also been recently proposed by \cite{lapa_increasing_2024}. This method internally applies the Wang-Mendel algorithm.
	
	
	Nonetheless, it can be reasonable to generate rules automatically, since the above list sure contains implausible rules, whose discovery as significant parts of the regression analysis would be strong indications of errors in the training data already. However, this method is recommended only as a secondary resort, if the search for suspected patterns based on manual rule definition fails.
\end{description}

The two methods have their individual use-cases:
\begin{itemize}
	\item If we search for potential inconsistencies within the training data, then an \emph{automatic} rule generation may help.
	\item If we search for known unwanted patterns, those would be defined \emph{manually}.
\end{itemize}

\subsection{Fitting the Regression Model}\label{sec:fitting}
The actual fitting of the model is based on the design matrix for the regression model
\begin{equation}\label{eqn:regression-model}
	r(\vec x)=\beta_0 + \beta_1\cdot f_1(\vec x)+\beta_2\cdot f_2(\vec x)+\ldots+ \beta_{m-1} \cdot f_{m-1}(\vec x)
\end{equation}
to minimize the sum of quadratic residuals $\sum_{i=1}^n (r(\vec x_i)-y_i)^2$, given $n$ training records $(\vec x_i,y_i)$ in the data matrix $\vec D$.

Adopting the exposition from \cite{press_numerical_1992}, given data points $\vec x_1,\ldots,\vec x_n$, and letting $x_{i,j}$ for $1\leq i\leq n$ and $1\leq j\leq k$ denote a single data item, the design matrix is found by evaluating the basis functions accordingly on all data points and arranging them in a matrix $\vec M$ as
\begin{equation}\label{eqn:design-matrix}
\vec M = \left(\begin{array}{ccccc}
	1 & f_1(\vec x_1) & f_2(\vec x_1) & \ldots & f_{m-1}(\vec x_1) \\
	1 & f_1(\vec x_2) & f_2(\vec x_2) & \ldots & f_{m-1}(\vec x_2) \\
	1 & \vdots & \vdots & \ddots & \vdots \\
	1 & f_1(\vec x_n) & f_2(\vec x_n) & \ldots & f_{m-1}(\vec x_n)
\end{array}\right)\in\R^{n\times m},
\end{equation}

from which the least-squares solution to the fitting problem gives the $\beta$-coefficients by multiplying the response vector (crisp values, not fuzzy memberships) with the pseudo-inverse
\[
(\vec M^\top\cdot \vec M)^{-1}\cdot \vec M^\top\cdot \vec y,
\]
in which the matrix $\vec M^\top\cdot \vec M\in\R^{m\times m}$ is presumed with full rank, which follows if the design matrix $\vec M$ has full column-rank (which in turn would require $n\geq m$ as a necessary condition).

If the number $m$ of rules is large, we may find $\text{rank}(\vec M^\top\cdot \vec M)<m$, which means duplicate or linearly dependent columns, i.e., rules that are in a way ``redundant'' (for example, if we have two rules, one of which says ``...\textbf{then} X is \emph{high}'' and the second saying under the same condition that ``...\textbf{then} X is \emph{very high}''; these rules would be candidates for linear dependence, or the matrix $\vec M^\top\vec M$ is at least badly conditioned).

To avoid this issue, we fit the model by a LASSO estimator, which (in its Lagrangian form) adds a penalty term $\lambda\cdot \norm{(\beta_0,\ldots,\beta_{m-1})}_1$ to simultaneously accomplish regularization and model size reduction. The caveat is that the usual significance testing of the model components becomes more involved \cite{lockhart_significance_2014} by requiring individual LASSO fits for leaving out each basis function $f_i$ once, where $i=1,2,\ldots,m-1$. Experimentally, this turned out to be relatively costly and hence inefficient. A simpler alternative method to first use the LASSO for a selection of rules towards a subset $F\subset\set{f_1,\ldots,f_{m-1}}$, then doing a regular least-squares fit with the basis functions from $F$ (only) and significance testing them (by standard means) either produced very few or very many significant basis functions, and hence led to strong under- or overfitting. Thus, a first LASSO-fit was performed to filter out rules with near-zero coefficients (threshold of $10^{-8}$). 


\subsection{Significance Determination}\label{sec:sig-testing}
Subsequently, a statistical test that de-biases the LASSO coefficients to compare them against a Gaussian distribution was applied \cite{geer_asymptotically_2014}, where no matrix inverse is taken, as nodewise LASSO is used for de-biasing, so cases where $m \geq n$ are not problematic. 

In essence, this resembles the standard method of including versus excluding a variable, in our case an if-then rule, from the model, and testing whether the model's explanatory quality is the same with or without the rule (null-hypothesis). If the statistical test rejects, we consider the respective rule as significant (to the user-supplied confidence level, which defaults to $\alpha=0.05$). The power of this test is to be considered separately (not further done in this work).


To account for many comparisons for countless automatically generated rules, the significance level $\alpha$ should be adjusted accordingly with the number of rules that were not filtered out from the near-zero-coefficient filter or one should at least be aware of this additional risk. For instance, using methods like Bonferroni correction \cite{miller2012simultaneous} will help in reducing the likelihood of falsely identifying a rule as significant.


\subsection{Tracing Rules Back to the Data Records}
\label{sec:rule-tracing}
Now, having found the rules that (significantly) approximate the given data $\vec D$, suppose that we actually found the records that manifest the unwanted pattern. The natural question from here is how to trace the rule back to the specific record that we may consider for removal or correction.

To this end, again standard procedures like leave-one-out cross-validations could help (e.g., testing whether the rule would remain significant if we were to remove single records), however, the removal of a single record may not change the significance of rules noticeably.  Furthermore, the computational load of such a procedure is immense for large $n$.

Instead, we propose the following heuristic to determine which records made a certain (unwanted) rule significant. The intuition is based on measuring how much a certain rule ``contributes'' to the actual model's predicted value. 

Given the final model in terms of if-then rules (as continuous functions from fuzzy expert systems), we can evaluate the sub-term $\beta_j\cdot f_j(\vec x_i)$ of the $j$-th rule set on the $i$-th data record $\vec x_i$, to determine its overall contribution to the model output. That is, with the same indexing as for the design matrix \eqref{eqn:design-matrix}, we iterate over the training records $\vec x_1,\ldots,\vec x_n$, and compute the following ratio:
\begin{equation}\label{eqn:heuristic-importance}
	\rho_{i,j}=\frac{{\abs{\beta_j\cdot f_j(\vec x_i)}}}{\sum_{k=1}^m \abs{\beta_k\cdot f_k(\vec x_i)}}\in[0,1]\quad\text{for~}j=1,2,\ldots,m
\end{equation}

This quantity gives an indication as to how much the rule $f_j$ ``adds'' to the final model outcome for record no. $i$, intentionally disregarding here the signs of the actual coefficient or the model's value. Records $x_i$ with a high value of $\rho_{i,j}$ have their label strongly influenced by rule $f_j$, hence can be suspected to contribute to the pattern that this rule embodies. While \eqref{eqn:heuristic-importance} is only a heuristic and has no statistical foundation, it has an intuitive interpretation as the ``relative contribution of a rule towards the model's prediction''. In other words, \eqref{eqn:heuristic-importance} measures how strong a record makes a rule fire, and records that make rules fire that encode unwanted patterns are candidates for closer inspection towards data cleaning or correction, e.g., if we suspect the training data to have been poisoned \cite{cina_wild_2023}.


Our implementation delivers the top ten records for the $j$-th rule by selecting the ten highest indices $i$ of of $\rho_{i,j}$, sorted in descending order. Thus, we normalize over all rules but choose the influence over all records, for a subsequent revision (repair) of the data matrix $\vec D$.


\section{Implementation as an Open Source Prototype}
Our prototype implementation is an open source GitHub project under GPL-3 license \cite{dallinger_s0urc10udxai-fuzzy-regrules_2024}. For the remainder of this document, we will use the latest version of the 20th November 2024 (commit hash \texttt{877fffe}) to describe the implementation and provide results. After cloning and compiling, a local \texttt{http} server can be started that lets the user interface be opened in a webbrowser at \texttt{localhost:8080}. Screenshots hereafter are taken from this graphical user interface in the browser.

\subsection{Vocabulary and Rule Definition}\label{sec:linguistic-variable-def}


The implementation uses triangular membership functions across a user-defined number of equidistant regions within the numerical range of the input parameters. The partition is made so that at most three triangles overlap towards unit sum; see Figure \ref{fig:example-linguistic-variable} for an illustration. These partitions are mapped to the categorical values in the dataframe. By default, we have three such categories (\emph{low} $<$ \emph{medium} $<$ \emph{high}), which are created by a division of the numeric range into three sub-intervals that overlap with the respective directly lower and directly upper category (see Figure \ref{fig:example-linguistic-variable}). 

The user can choose up to seven equidistant categories (\emph{very low} $<$ \emph{low} $<$ \emph{medium low} $<$ \emph{medium} $<$ \emph{medium high} $<$ \emph{high} $<$ \emph{very high}) for fuzzification and independently for the defuzzification of the target variable. While increasing the number of categories improves precision, it also introduces more linear dependence due to overlapping membership functions during fuzzification and the activation weighting process in defuzzification. The defuzzification process uses the Middle of Maxima method \cite{gilda_2022_Defuzzification}, scaled by the total activation degree of the antecedents, similarly contributing to linear dependence. Nevertheless, the use of more categories allows for a finer-grained data representation.


\subsection{Model Preparation and Rule Reduction}\label{sec:rule-cleaning}
Once a comma separated value (CSV) file is uploaded, the model is automatically set up by:
\begin{enumerate}
	\item constructing linguistic variables for each field in the loaded data frame (CSV file), with the number of categories (for continuous variables) as chosen by the user, and other (nominal) categories defined from the values found in the file. Linguistic variables and fuzzifiers are constructed as Sections \ref{sec:automated-rule-generation} and \ref{sec:linguistic-variable-def} described.
	
	The interface provides a whitelist and blacklist for manual rule inclusion or exclusion, where users can enter one rule per line (Figure \ref{fig:bw-list}). Rules for exclusion from the automatic generation go on the blacklist, while rules intended to explain the data -- also including undesired patterns -- can be added to the whitelist for significance testing.
	
	\begin{figure}[h!]
		\centering
		\includegraphics[scale=0.15]{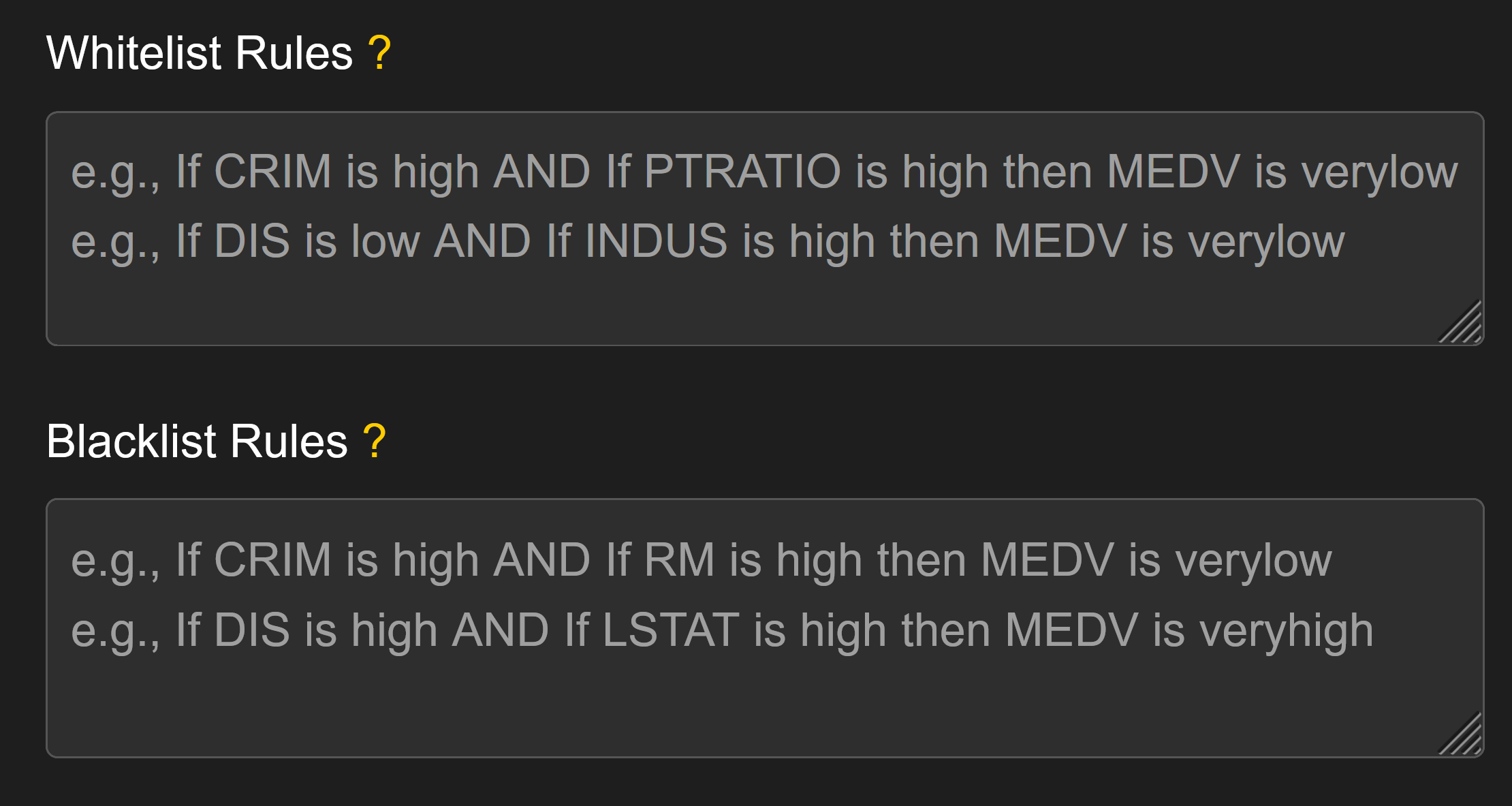}
		\caption{Interface to black- and whitelist rules explicitly}\label{fig:bw-list}
	\end{figure}
	
	We emphasize that users can set the maximum number of antecedents for rule generation. However, to prevent exponential growth of the number of rules, potential overfitting and very long computation time, we recommend keeping the number of AND-ed antecedents as low as possible but as high as required to explain the dataset (i.e., to get reasonable performance metrics). By default, we auto-generate rules with up to two-variable antecedents, while more complex rules can be added via the whitelist. By setting a limit to the number of antecedents, the exponential number of possible rules is reduced to polynomially many (of a degree being the number of allowed antecedents). Our implementation creates one basis function per if-then rule.
	
	\item fitting the model and significance testing: This works along the lines of Section \ref{sec:fitting}, including an automatic rule cleaning using a LASSO fit, and significance testing according to \cite{geer_asymptotically_2014}. A flag on the user interface lets insignificant rules automatically disappear from the output list if desired (even if they came from the whitelist). Figure \ref{fig:biased-salaries-example} shows an example of the output, where the column ``Coefficient'' represents $\beta_j$ for rule $f_j$, and the ``Trend'' column is a graphical representation of the consequent to provide a simple method for filtering and searching.
	
	In addition to the LASSO fit, we also drop rules under the following conditions in the respective order:
	\begin{itemize}
		\item The rule priority falls under a certain user-specifiable threshold, where the rule prioritization is based on known heuristics from machine learning \cite{aggarwal_2015_DataMT}. Specifically, we model the priority as a weighted linear combination of the following factors (which are displayed in Figure \ref{fig:biased-salaries-example} as part of the output), where the user can set the weight for each factor (this value does not, however, affect or determine the respective $\beta$-coefficient in \eqref{eqn:regression-model}):
		\begin{itemize}
			\item the \emph{support} \cite{agrawal_1994_fast} of a rule $A\to B$ (abbreviating an ``\textbf{if} $A$ \textbf{then} $B$'' statement). This is the fraction of records on which a rule would fire, if the data was discretized in linguistic terms that the rules use, and then treated as for crisp logic. Technically, we map all continuous values to a (fuzzy) category from $\{$\emph{very low}, \ldots, \emph{very high}$\}$ of highest membership to discretize the values. The support is then the fraction of records for which both the rule's antecedent and consequent are true, i.e., $\supp(A\to B)=\frac 1 n\cdot (\text{number of records with predicates $A$ and $B$ true})$. Intuitively, this gives a more useful indication of how often a rule comes to fire, since the actual fuzzy inference would naturally let a rule almost always apply if the antecedent's membership evaluates to anything $>0$ (no matter how small).
			
			\item the \emph{leverage} \cite{bouchon_2011_modern} of a rule $A\to B$, which is $\supp(A\to B)-\supp(A)\cdot \supp(B)$. This value measures independence in the same way as it were quantifiable for random variables by the value $\Pr(X|Y)-\Pr(X)\cdot \Pr(Y)$, which is zero if and only if the events $X$ and $Y$ are stochastically independent.
			\item the \emph{number of antecedent predicates}, as well as an additional weight that the rule may get from being on the \emph{whitelist}.
		\end{itemize}
		The larger the above values are for a rule, the higher the priority (the above scores can additionally be weighted in a user-definable way). All basis functions are sorted according to priority, meaning that for instance in the following duplicate column removal, the higher prioritized rule survives. Note that the said discretization was only created to obtain the support and the leverage, and is thereafter again discarded. 
		\item The rule creates a duplicate and hence redundant column in the design matrix; since exact equality of columns has zero probability, we apply a likewise roundoff threshold (user-definable) as to when to consider a column as ``redundant'' and hence for removal. The implementation records all such deletions and lists the removed rules as secondary (meaning the basis function is equivalent with a second linguistic meaning over all records).
		\item Their LASSO coefficient falls under a certain threshold of acceptance (by default $10^{-8})$, in which case the numeric contribution of a rule would likely be too small to notice or even vanish in a roundoff error.
	\end{itemize}
	Removed rules are -- together with the respective reason -- printed in the warnings-section of the output. Furthermore, the goodness of fit is displayed in a separate section below the rule results.
\end{enumerate}





\section{Example Datasets and Results}\label{sec:evaluation}

We provide several datasets to validate the method and demonstrate the system's capabilities.
\subsection{The Sanity Check Data}\label{sec:sanity-check}

The \emph{sanity check} dataset is artificial and intended to verify if the method can detect a purely random variable as insignificant, in comparison to a value that deterministically depends on some other variable. To this end, the dataset contains some columns filled with uniform random values, and one ``data'' column with real values (floating point numbers), where the (fuzzy) categories \emph{low}, \emph{medium}, etc. are mapped over (i.e., the data columns map to the respective discretized terms directly via the identity function). This dataset shows that the system can indeed distinguish between random and deterministic variables, as the generated rules containing random variables are not found significant, while the rules corresponding to the data generation process are. Furthermore, the rules which correspond to the data generation process have at least ten times higher coefficients than the rules generated from random effects.

\subsection{The Boston Housing Dataset \cite{perera_boston_2018}}

Analyzing this dataset with the median value of owner-occupied homes in US-Dollar 1000's (variable ``\texttt{MEDV}'' in the dataset) as a target, we found a total of 433 rules explaining this data significantly under the default configuration of the software. The fuzzy rules indeed fit the data quite well (for example, with a mean absolute percentage error of $\approx 3.81\%$, so the accuracy is quite in the range of contemporary machine learning models). Given a reasonable accuracy, the main purpose of the model is the discovery of interesting, unexpected, or even unwanted patterns in the data. However, as we will discuss later in Section \ref{sec:discussion}, some explanations for a dataset can be less direct than others and hide true causes, but still remain true.

Deviating from the default configuration by using only one antecedent (so significantly fewer automatically generated rules), five fuzzy sets for (de-)fuzzi\-fication (additionally including \emph{very low} and \emph{very high}) and reducing regularization ($\lambda$ = 0.1) resulted in a higher mean absolute percentage error of 10\%, but yielded much more directly intriguing and comparable results for this dataset. The following -- from our perspective interesting -- rules were identified:

\begin{table}[h]
	\scriptsize
	\centering
	\caption{A subset of fuzzy rules discovered to explain the Boston Housing dataset}
\label{tab:significant_rules}
	\begin{tabular}{|l|c|c|c|c|}
	\hline
	\textbf{Rule} & \textbf{Coeff.} & \textbf{Support} & \textbf{Leverage} & \textbf{p-value} \\
	\hline
	\textbf{If} \texttt{LSTAT} is \emph{high} \textbf{then} \texttt{MEDV} is \emph{low} & 2.91 & 0.0277 & -0.00175 & 0.5286 \\
	\textbf{If} \texttt{B} is \emph{high} \textbf{then} \texttt{MEDV} is \emph{very low} & 0.96 & 0.0099 & 0.0058 & 0.0245 \\
	\textbf{If} \texttt{NOX} is \emph{medium} \textbf{then} \texttt{MEDV} is \emph{low} & 9.29 & 0.1403 & 0.0236 & $1.12 \times 10^{-12}$ \\
	\textbf{If} \texttt{NOX} is \emph{medium} \textbf{then} \texttt{MEDV} is \emph{very high} & 1.50 & 0.0237 & 0.0127 & $5.34 \times 10^{-6}$ \\
	\textbf{If} \texttt{RM} is \emph{very high} \textbf{then} \texttt{MEDV} is \emph{high} & 9.07 & 0.00395 & 0.00255 & $1.78 \times 10^{-15}$ \\
	\textbf{If} \texttt{CRIM} is \emph{very high} \textbf{then} \texttt{MEDV} is \emph{very low} & 0.92 & 0.00198 & 0.00186 & 0.0254 \\
	\textbf{If} \texttt{TAX} is \emph{very low} \textbf{then} \texttt{MEDV} is \emph{very high} & 0.90 & 0.01186 & 0.00548 & 0.0061 \\
	\hline
	\end{tabular}
	\normalsize
\end{table}

Among the shown rules in Table \ref{tab:significant_rules}, the first rule, where \texttt{LSTAT} represents the percentage of lower status of the population within the town, aligns with the intuitive understanding that a higher proportion of ``lower-status'' individuals in a neighborhood is associated with lower median home values. However, the high p-value of 0.5286 indicates that this rule is not statistically significant, suggesting that the relationship may not be robust when accounting for variability in the data. This lack of significance is likely due to large standard errors associated with the basis function for \texttt{LSTAT}, which diminishes the confidence in this association despite its logical consistency.

The second rule, \textbf{If} \texttt{B} is \emph{high} \textbf{then} \texttt{MEDV} is \emph{very low}, reveals a statistically significant pattern. The variable \texttt{B} represents the proportion of black residents in the town. This rule indicates that high proportions of black individuals are associated with lower median home values. While the coefficient for this rule is lower (0.96) compared to the first rule, its statistical significance highlights a concerning and unethical pattern of racial discrimination embedded within the dataset. This finding underscores the importance of scrutinizing models for biased or discriminatory patterns that may inadvertently reinforce societal inequalities.

As a sanity check, we can see that \texttt{NOX} (the nitric oxides concentration) has vast influence on the housing prices as the rules with \texttt{NOX} have notably higher coefficients, which was presumed by the authors of the dataset. Notably, conflicting rules were also identified, revealing a complex relationship between \texttt{NOX} levels and median home values, as is shown in the second and third row of Table \ref{tab:significant_rules}.

These conflicting rules suggest that while medium levels of \texttt{NOX} are generally associated with lower housing prices, there are exceptions where medium \texttt{NOX} levels correspond to very high median values, indicating potential underlying factors or interactions not immediately apparent.

\subsection{The Biased Salaries Dataset}\label{sec:biased-salaries-example}
This dataset was created artificially to bring in effects or patterns, and to test if the method would re-discover these patterns without being a priori informed about them. The data itself was created using a Python script, which sets salaries for employees according to individual (un-)ethical rules, like a preference of certain human resource managers for the third gender (unethical preference over other genders), increasing salaries with grade point average (GPA), university reputation and experience (ethically admissible), but presuming women with generally less experience (unethical pattern), which then indirectly causes less salary for this group.

\begin{figure}
		\centering
		\includegraphics[width=\textwidth]{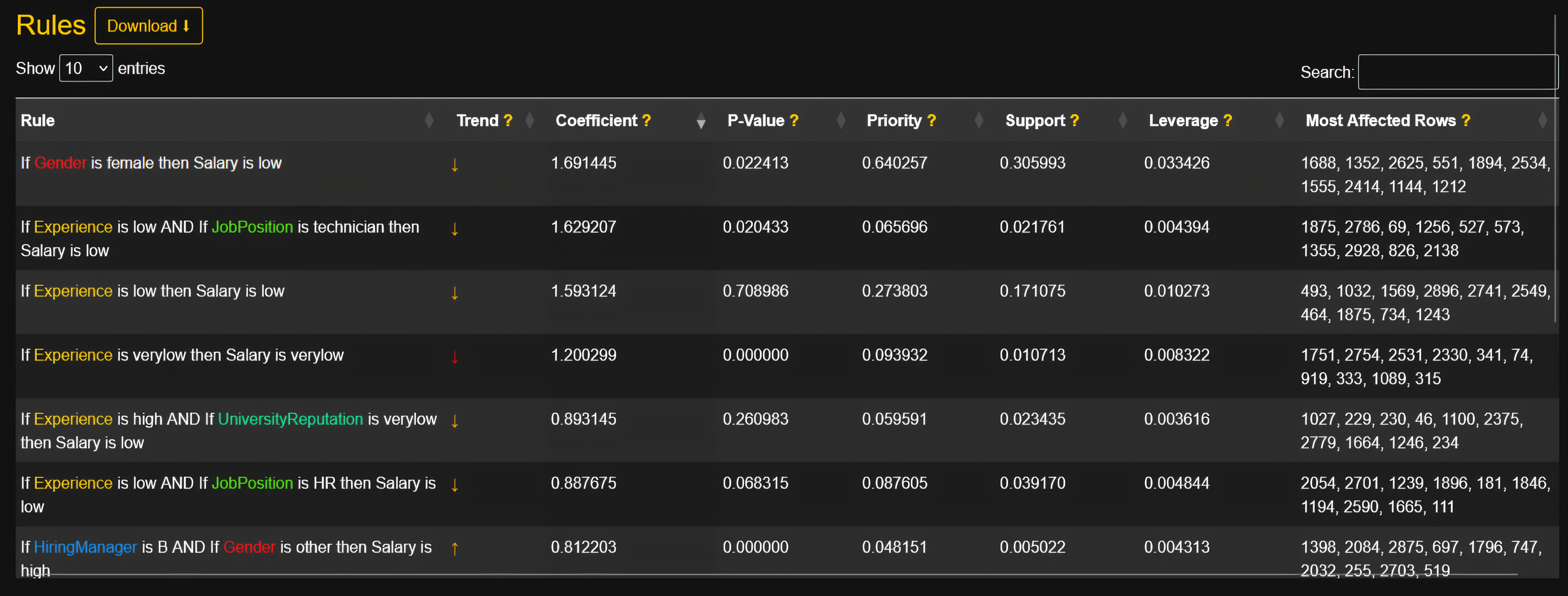}
		\caption{Results from the demo dataset \cite{dallinger2024} with unethically decreased salaries for women; the rule ``\textbf{if} Gender is \emph{female} \textbf{then} Salary is \emph{low}'' is found significant to explain this data; (used parameters: $\lambda=0.1$, maximum LASSO iterations: $1000$)}
		\label{fig:biased-salaries-example}
	\end{figure}

The dataset was loaded into the system, letting it automatically create rules that exhaustively combine all variables and their possible values. This list would naturally contain admissible, but also unethical heuristics to explain the data. Our method's primary purpose is to distill a reasonably small fraction of this large rule set, to explain the data well, and thereby point more directly to unethical patterns in the training data. Indeed, the fact that women earn less is re-discovered; even despite the fact that it is an \emph{indirect} consequence of the mechanism that created the data (the salary reduction is due to less experience that the gender implies, so the gender does not ``directly'' cause this reduction). Figure \ref{fig:biased-salaries-example} shows the outcome, where -- for a deeper inspection -- gender-related rules can be directly filtered by using the ``Search'' textbox above the rules.

	
\section{Discussion}\label{sec:discussion}

The rules generated to explain the data can, subsequently, be treated as crisp logic and be checked for logical consistency. If the entire set of rules is identified with inner logical inconsistencies, then a standard model-based diagnosis \cite{reiter_theory_1987} is possible to identify the smallest set of rules to be removed from the model to regain consistency. However, the usual removal-and-consistency checking method of model-based diagnosis may be undesirable, since the model already \emph{is minimal} in the sense of containing only rules that are statistically significant, meaning that any further removal would degrade the accuracy. 

\begin{figure}
	\centering
	\includegraphics[width=\textwidth]{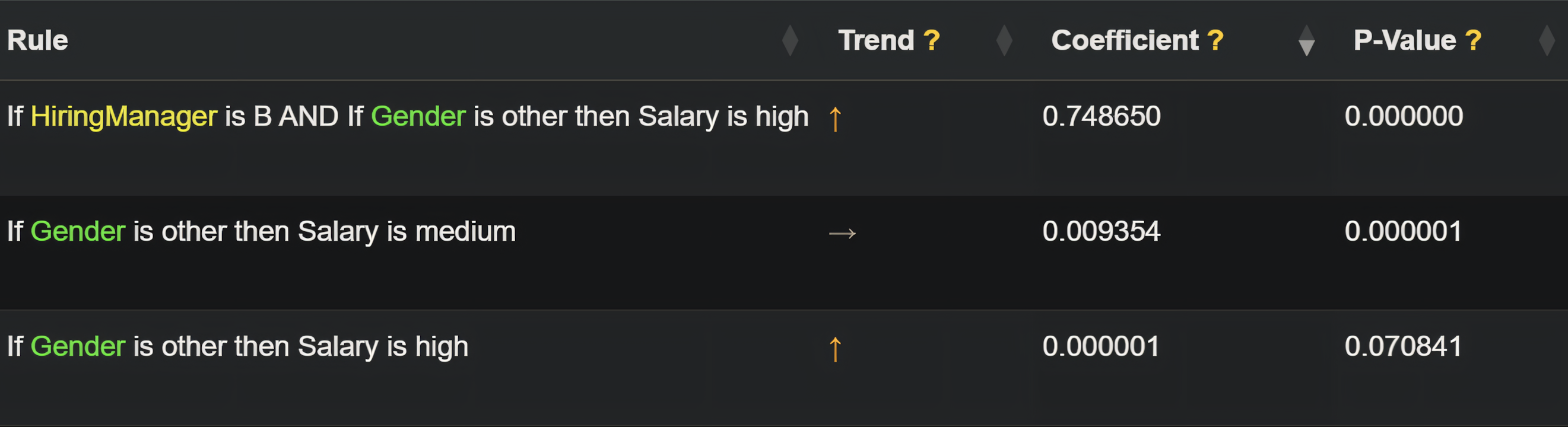}
	\caption{Exemplary inconsistent rules found in the biased salaries dataset (with $\lambda=50$ and maximum LASSO iterations: $10000$)}
	\label{fig:inconsistent-rules}
\end{figure}

Consequently, we may be less interested in a (logic-based) diagnosis, but rather in the conflict sets within the rules, i.e., the smallest sets of rules whose removal would resolve the inconsistency. Conflict sets are much faster computable \cite{junker_quickxplain:_2004}, and can be interpreted as descriptions of undesired patterns within the data. Logical reasoning to decide consistency or inconsistency, however, needs to account also for the correlation of the rule with the data itself. As mentioned before, negative coefficients of a rule (see below for examples) may require revising rules before checking their consistency against others. Our own implementation is (currently) a bit simpler here in using a Python script\footnote{\label{ftn:logical-checkup-script}\cite[\texttt{/example\_unveiling\_biases/simple\_contradiction\_check.py}]{dallinger_s0urc10udxai-fuzzy-regrules_2024}} to scan for pairs of rules with logically inconsistent statements. The script identifies directly ``conflicting rules'' where precisely identical conditions (antecedents) lead to different conclusions (consequents), as well as ``specializing rules'' where a more specific rule has a different consequent compared to the general rule, thus revealing patterns where added conditions influence the target variable inconsistently.
For the example of the biased salaries dataset, such inconsistencies were indeed found, namely there were records that assigned medium salary to people of the third gender, while (under stronger conditions), people from the same group would receive high salaries; see Figure \ref{fig:inconsistent-rules}. This is an example of a logically inconsistent pair of rules, which is easy to detect, but points out a (possibly unfair) positive bias of some hiring managers towards people of non-binary gender (again, in our artificial dataset only).

The hyperparameter for the LASSO regularization (the value of $\lambda$) has a strong impact on which rules are found; for example, in the demo dataset with the biased salaries, we may or may not directly obtain the rule that ``women earn less'', depending on the setting of the hyperparameter and how long the iterative LASSO descent is performed (i.e. how close one gets to the fixed point). However, the discovery of women earning less remained intact and present in all our experiments, only with the difference of this unethical finding being pointed out concisely (by a direct rule) or rather being implied by several other rules from which the same discovery (``women earn less'') would follow. As a provocative analogy: mathematical truths remain the same, whether we only consider basic axioms or advanced theorems; the latter are only more specific on the important facts.


The direct observation that women earn less was frequently identified when the LASSO regularization terminated earlier (i.e., farther from the fixed point). This phenomenon might be linked to overfitting, where a more abstract idea is overly specified by incorporating numerous complex or concrete factors (often even abusing noise). As iterations increase, these ideas tend to become more concrete. 

Furthermore, the overall phenomenon of being heavily reliant on the hyperparameters of the regularization may be closely linked to the findings of Pantazis et al. \cite{pantazis_2018_multipleequivalentsolutionslasso}, who demonstrate that LASSO can produce multiple solutions when perfect multicollinearity is present, while still remaining a convex problem because of the least squares performance metric in this case. The multicolinearity arises in our case due to the way we (de-)fuzzified and generated the rules. 

Generally, if logically inconsistent rules are found, it points at data patterns that apparently follow a flawed, or at least mutually incompatible, logic. Unless this logic is hypothesized a priori, an automatic rule generation could become desirable, since the ``blind'' generation of all possible rules will naturally also produce logically inconsistent rules, which should not be simultaneously significant. 
In a further experiment, we modified the biased salaries dataset by replacing all salaries of men with random samples from a Weibull distribution, while letting the females and other employees receive (random) salaries sampled from a Gamma distribution. The parameters of both distributions were chosen to let them have identical means and variances, so that a statistical test about equality of the average income between the genders (male vs. ``female or other'') would not detect any discrepancies. However, the distributions have quite different skews, so that a residual discrepancy in the salaries \emph{does} remain. Analyzing this dataset with the system using automatically generated rules brought up a large set of logically inconsistent pairs, already among rules that include the gender of a person. Among the rules found (with the configuration $\lambda=1, \text{maximum number of iterations}=1000$) are the following, with respective coefficients and p-values added in brackets:
\begin{quote}
	\textbf{if} GPA is \emph{very high} \textbf{AND} Gender is \emph{male} \textbf{then} Salary is \emph{high}\\($\beta_i = 0.0746$, $p_i = 0.4647$)

	\textbf{if} GPA is \emph{very high} \textbf{AND} Gender is \emph{male} \textbf{then} Salary is \emph{medium}\\($\beta_i = -0.0769$, $p_i = 0.9274$)
	
	\textbf{if} GPA is \emph{very high} \textbf{AND} Gender is \emph{male} \textbf{then} Salary is \emph{very low}\\($\beta_i = 0.0449$, $p_i = 0.2333$)
\end{quote}

The high $p$-values across these rules indicate insufficient statistical significance at any reasonable threshold, which aligns with the observation that salary distribution depends solely on gender in this context. However, the non-zero coefficients surviving initial LASSO filtering allow these rules to serve as a demonstration for the hypothetical case that they could be found significant in some other dataset. While the rules appear contradictory (and logically are), their coefficients' opposing signs in the linear model reveal how their combined effects influence salary predictions. Specifically, the negative correlation of the second rule suggests it may be better interpreted as predicting deviations from medium salaries, aligning with actual data trends where such predictions are negatively correlated (as the chosen salary distribution is not symmetric).

We therefore highlight that logical inconsistencies judged without considering coefficient signs can be misleading. The diagnostic script (see Footnote \ref{ftn:logical-checkup-script}) accounts for this by filtering rules based on user-defined $\beta$-thresholds (defaulting to zero) and significance levels. In this simulation, where GPA lacks influence, these outcomes might reflect random effects, but significant rules would expose true contradictions, such as the incompatibility between the first and last rules. Indeed, however, we want to remark that this discrepancy could be a side-effect of the non-symmetric distribution.

Quite consistently with how the salaries were chosen in this experiment, the following rules were also found (among many others that we omit here for brevity):
\begin{quote}
	\textbf{if} Gender is \emph{female} \textbf{then} Salary is \emph{very low} ($\beta_i = 0.0039$, $p_i = 0.1559$)
	
	\textbf{if} Gender is \emph{female} \textbf{then} Salary is \emph{medium} ($\beta_i = 0.0027$, $p_i = 0.3737$)
	
	\textbf{if} Gender is \emph{female} \textbf{then} Salary is \emph{low} ($\beta_i = 0.0009$, $p_i = 0.1913$)
	
	\textbf{if} Gender is \emph{female} \textbf{then} Salary is \emph{very high} ($\beta_i = -0.0001$, $p_i = 0.1581$)
	
	\textbf{if} Gender is \emph{male} \textbf{then} Salary is \emph{low} ($\beta_i = -0.0185$, $p_i = 0.8686$)
	
	\textbf{if} Gender is \emph{male} \textbf{then} Salary is \emph{very low} ($\beta_i = -0.0231$, $p_i = 0.8415$)
	
	\textbf{if} Gender is \emph{male} \textbf{then} Salary is \emph{medium} ($\beta_i = -0.0687$, $p_i = \mathbf{0.0088}$, significant!)
	
	\textbf{if} Gender is \emph{male} \textbf{then} Salary is \emph{high} ($\beta_i = 0.1084$, $p_i = 0.1653$)
\end{quote}

The generated rules display logical inconsistencies, such as conflicting salary predictions for the same gender under similar conditions, which may initially suggest issues with the model's coherence. These inconsistencies, however, likely arise from the (differing) skewness of the Weibull and Gamma distributions used to generate male and female salaries, despite having identical means and variances. Therefore, the model's gender-specific rule conflicts are artifacts of the underlying data distribution shapes rather than evidence of inherent bias. They still point to a hidden issue with the data, which was here artificially injected, but we provided no prior information on this when the analysis started.
 Alternatively, such inconsistent rules may also exist due to badly defined linguistic variables or too small rulesets. This is why sanity checks, like in Section \ref{sec:sanity-check}, but on an artificially created dataset for testing and validation purposes, are recommendable as a prior step.


Furthermore, the negative sign of the coefficient for the rule, indicating that the consequent ``Salary is medium'' for males negatively correlates with the data (which is also statistically significant), indicates that there \emph{is a difference} between the two groups (which a statistical test would have missed, since the data was dredged accordingly towards tricking the test). We stress that the example was here made with a dataset that additionally violates the distributional assumptions on the error term (since the variables have skewed distributions). The results of the fit are hence generally not optimal and a generalized linear model would be appropriate. Nonetheless, the indication of an underlying issue with the data is still reflected in the rules distilled, even though the exact root cause of the problem does not become ultimately clear (due to the way the data is constructed).

Moreover, to gain a comprehensive understanding of the rules underlying a dataset, it might be advantageous to designate several fields in the data as dependent variables to be explained by the other fields. In the previous setting where we worked on the original biased salaries dataset with our software, we only explained one (target) field, but considering the generated rule ``\textbf{if} Experience is \emph{low} \textbf{then} Salary is \emph{low}'', it could also be beneficial to try to explain the experience field as only then one can obtain ``\textbf{if} Gender is \emph{female} \textbf{then} Experience is \emph{low}''. In our experimentation, however, we noticed that ``\textbf{if} Gender is \emph{female} \textbf{then} Salary is \emph{low}'' was directly obtained for certain regularization parameters. 

\section{Conclusion and Outlook}
The system is currently implemented as a proof of concept to validate the theoretical ideas laid out in the text. The demonstration system serves verification purposes only, and is designed for ease of use, but not to explore the full degrees of freedom that the fuzzy modeling would offer, nor with any built-in hyperparameter tuning. That is, future versions and work may include support for type-2 fuzzy sets, rules with more antecedents, sensitivity analyses of how much the model is affected by changing the underlying heuristics to convert Boolean expressions into continuous functions (for example, when we do not use min-max operators for the Boolean AND/OR, but a general $t$-norm, or using other implications like $(p\to q)\Leftrightarrow (\neg p\lor q)$ converted into $\max(1-\mu_p, \mu_q)$ as a membership function). The automatic rule generation could use orthogonal arrays \cite{kampel_survey_2019} to escape combinatorial explosions, and algorithms from Reiter's model based diagnosis \cite{reiter_theory_1987} and conflict set computation \cite{junker_quickxplain:_2004} can be brought in. However, any reasoning-based sanity check of the final rules must take into account that the fuzzy nature of rules makes an ``if-then'' more a correlation than a deduction towards the data \cite{mendel_2021_critical}; hence the question whether mutually contradictive rules also have a logically contradictive meaning is a question of future research interest. 

	

\begin{acks}
This work has been supported by the LIT Secure and Correct Systems Lab funded by the State of Upper Austria and the Linz Institute of Technology (LIT-2019-7-INC-316).
\end{acks}

\bibliographystyle{abbrv}
\bibliography{fuzzy-xai_arxiv}

\begin{thebibliography}{10}

\bibitem{aggarwal_2015_DataMT}
C.~C. Aggarwal.
\newblock {\em Data Mining: The Textbook}.
\newblock Springer Publishing Company, Incorporated, 2015.

\bibitem{agrawal_1994_fast}
R.~Agrawal and R.~Srikant.
\newblock Fast algorithms for mining association rules in large databases.
\newblock In {\em Proceedings of the 20th International Conference on Very
  Large Data Bases}, VLDB '94, pages 487--499, San Francisco, CA, USA, 1994.
  Morgan Kaufmann Publishers Inc.

\bibitem{AshokanH21}
A.~Ashokan and C.~Haas.
\newblock Fairness metrics and bias mitigation strategies for rating
  predictions.
\newblock {\em IP\&M}, (5), 2021.

\bibitem{Baeza_Yates18}
R.~Baeza{-}Yates.
\newblock {Bias on the Web}.
\newblock {\em Communication {ACM}}, 61(6):54--61, 2018.

\bibitem{barocas_2023_fairness}
S.~Barocas, M.~Hardt, and A.~Narayanan.
\newblock {\em Fairness and Machine Learning: Limitations and Opportunities}.
\newblock Adaptive Computation and Machine Learning series. MIT Press, 2023.

\bibitem{becker_discrimination_2019}
S.~O. Becker, A.~Fernandes, and D.~Weichselbaumer.
\newblock {Discrimination} in hiring based on potential and realized fertility:
  {Evidence} from a large-scale field experiment.
\newblock {\em Labour Economics}, 59:139--152, Aug. 2019.
\newblock online:
  \href{https://www.sciencedirect.com/science/article/pii/S0927537119300429}{https://www.sciencedirect.com/science/article/pii/S0927537119300429},
  DOI:
  \href{https://www.doi.org/10.1016/j.labeco.2019.04.009}{10.1016/j.labeco.2019.04.009}
  [retrieved: 2024-11-04].

\bibitem{blodgett_llms_2024}
S.~L. Blodgett and Z.~Talat.
\newblock {LLMs} produce racist output when prompted in {African} {American}
  {English}.
\newblock {\em Nature}, 633(8028):40--41, Sept. 2024.
\newblock Bandiera\_abtest: a Cg\_type: News And Views Publisher: Nature
  Publishing Group Subject\_term: Machine learning, Language, Society, online:
  \href{https://www.nature.com/articles/d41586-024-02527-x}{https://www.nature.com/articles/d41586-024-02527-x},
  DOI:
  \href{https://www.doi.org/10.1038/d41586-024-02527-x}{10.1038/d41586-024-02527-x}
  [retrieved: 2024-11-04].

\bibitem{bollaert_frri_2025}
H.~Bollaert, M.~Palangetic, C.~Cornelis, S.~Greco, and R.~Slowinski.
\newblock {FRRI}: a novel algorithm for fuzzy-rough rule induction.
\newblock {\em Information Sciences}, 686:121362, Jan. 2025.
\newblock arXiv:2403.04447 [cs].

\bibitem{bonilla2006racism}
E.~Bonilla-Silva.
\newblock {\em Racism without racists: Color-blind racism and the persistence
  of racial inequality in the United States}.
\newblock Rowman \& Littlefield Publishers, 2006.

\bibitem{bouchon_2011_modern}
B.~Bouchon-Meunier, G.~Coletti, and R.~R. Yager.
\newblock {\em Modern Information Processing: From Theory to Applications}.
\newblock Elsevier Science Inc., USA, 2011.

\bibitem{chamola_review_2023}
V.~Chamola, V.~Hassija, A.~R. Sulthana, D.~Ghosh, D.~Dhingra, and B.~Sikdar.
\newblock A {Review} of {Trustworthy} and {Explainable} {Artificial}
  {Intelligence} ({XAI}).
\newblock {\em IEEE Access}, 11:78994--79015, 2023.

\bibitem{chimatapu_explainable_2018}
R.~Chimatapu, H.~Hagras, A.~Starkey, and G.~Owusu.
\newblock Explainable {AI} and {Fuzzy} {Logic} {Systems}.
\newblock In D.~Fagan, C.~Martín-Vide, M.~O'Neill, and M.~A. Vega-Rodríguez,
  editors, {\em Theory and {Practice} of {Natural} {Computing}}, pages 3--20,
  Cham, 2018. Springer International Publishing.

\bibitem{chukhrova_fuzzy_2019}
N.~Chukhrova and A.~Johannssen.
\newblock Fuzzy regression analysis: {Systematic} review and bibliography.
\newblock {\em Applied Soft Computing}, 84:105708, Nov. 2019.

\bibitem{cichy_fuzzy-approximation_2019}
C.~Cichy and S.~Rass.
\newblock {A} {Fuzzy-Approximation} {Approach} to {Explainable} {Data}
  {Quality} {Assessment}.
\newblock In {\em Proceedings of the 34th {International} {Business}
  {Information} {Management} {Association} {Conference} ({IBIMA})}, pages
  3919--3931, 2019.

\bibitem{cichy_overview_2019}
C.~Cichy and S.~Rass.
\newblock {An} {Overview} of {Data} {Quality} {Frameworks}.
\newblock {\em IEEE Access}, 7:24634--24648, 2019.
\newblock online:
  \href{https://ieeexplore.ieee.org/document/8642813/}{https://ieeexplore.ieee.org/document/8642813/},
  DOI:
  \href{https://www.doi.org/10.1109/ACCESS.2019.2899751}{10.1109/ACCESS.2019.2899751}.

\bibitem{cichy_fuzzy_2020}
C.~Cichy and S.~Rass.
\newblock {Fuzzy} {Expert} {Systems} for {Automated} {Data} {Quality}
  {Assessment} and {Improvement} {Processes}.
\newblock In {\em Proceedings of the {EKAW} 2020 {Posters} and {Demonstrations}
  {Session}}, volume 2751, pages 7--11. CEUR Workshop Proceedings, 2020.

\bibitem{cina_wild_2023}
A.~E. Cinà, K.~Grosse, A.~Demontis, S.~Vascon, W.~Zellinger, B.~A. Moser,
  A.~Oprea, B.~Biggio, M.~Pelillo, and F.~Roli.
\newblock Wild {Patterns} {Reloaded}: {A} {Survey} of {Machine} {Learning}
  {Security} against {Training} {Data} {Poisoning}.
\newblock {\em ACM Comput. Surv.}, 55(13s), July 2023.
\newblock Place: New York, NY, USA Publisher: Association for Computing
  Machinery.

\bibitem{comas_interval-valued_2025}
D.~S. Comas, G.~J. Meschino, and V.~L. Ballarin.
\newblock Interval-valued fuzzy predicates from labeled data: {An} approach to
  data classification and knowledge discovery.
\newblock {\em Information Sciences}, 707:122033, July 2025.

\bibitem{dallinger_s0urc10udxai-fuzzy-regrules_2024}
M.~Dallinger.
\newblock {S0urC10ud}/xai-fuzzy-regrules, Nov. 2024.
\newblock original-date: 2024-10-11T09:29:24Z.

\bibitem{dallinger2024}
M.~Dallinger.
\newblock xai-fuzzy-regrules/example\_unveiling\_biases/biased\_salaries at
  main · {S0urC10ud}/xai-fuzzy-regrules, Nov 2024.

\bibitem{death_2000_CARTAP}
G.~De'ath and K.~E. Fabricius.
\newblock Classification and regression trees: A powerful yet simple technique
  for ecological data analysis.
\newblock {\em Ecology}, 81(11):3178--3192, 2000.

\bibitem{Ekstrand0B022}
M.~D.~E. et~al.
\newblock Fairness in information access systems.
\newblock {\em Found. Trends IR}, 2022.

\bibitem{ferrara_fairness-aware_2024}
C.~Ferrara, G.~Sellitto, F.~Ferrucci, F.~Palomba, and A.~De~Lucia.
\newblock Fairness-aware machine learning engineering: how far are we?
\newblock {\em Empirical Software Engineering}, 29(1):9, Jan. 2024.

\bibitem{fumanal-idocin_ex-fuzzy_2024}
J.~Fumanal-Idocin and J.~Andreu-Perez.
\newblock Ex-{Fuzzy}: {A} library for symbolic explainable {AI} through fuzzy
  logic programming.
\newblock {\em Neurocomputing}, 599:128048, Sept. 2024.

\bibitem{geer_asymptotically_2014}
S.~v.~d. Geer, P.~Bühlmann, Y.~Ritov, and R.~Dezeure.
\newblock On asymptotically optimal confidence regions and tests for
  high-dimensional models.
\newblock {\em The Annals of Statistics}, 42(3):1166--1202, June 2014.
\newblock Publisher: Institute of Mathematical Statistics.

\bibitem{gilda_2022_Defuzzification}
K.~Gilda and S.~Satarkar.
\newblock Defuzzification: Maxima methods with improved efficiency and their
  performance evaluation.
\newblock In {\em 2022 IEEE Delhi Section Conference (DELCON)}, 02 2022.

\bibitem{gonzalez-sendino_mitigating_2024}
R.~González-Sendino, E.~Serrano, and J.~Bajo.
\newblock Mitigating bias in artificial intelligence: {Fair} data generation
  via causal models for transparent and explainable decision-making.
\newblock {\em Future Generation Computer Systems}, 155:384--401, June 2024.

\bibitem{hedayat_orthogonal_1999}
A.~S. Hedayat, N.~J.~A. Sloane, and J.~Stufken.
\newblock {\em Orthogonal arrays: theory and applications}.
\newblock Springer series in statistics. Springer, New York, 1999.

\bibitem{Heidari_2019}
A.~Heidari, J.~McGrath, I.~F. Ilyas, and T.~Rekatsinas.
\newblock Holodetect: Few-shot learning for error detection.
\newblock In {\em Proceedings of the 2019 International Conference on
  Management of Data}, SIGMOD/PODS ’19, page 829–846. ACM, June 2019.

\bibitem{hofmann_ai_2024}
V.~Hofmann, P.~R. Kalluri, D.~Jurafsky, and S.~King.
\newblock {AI} generates covertly racist decisions about people based on their
  dialect.
\newblock {\em Nature}, 633(8028):147--154, Sept. 2024.
\newblock Publisher: Nature Publishing Group, online:
  \href{https://www.nature.com/articles/s41586-024-07856-5}{https://www.nature.com/articles/s41586-024-07856-5},
  DOI:
  \href{https://www.doi.org/10.1038/s41586-024-07856-5}{10.1038/s41586-024-07856-5}
  [retrieved: 2024-11-04].

\bibitem{huhn_furia_2009}
J.~Hühn and E.~Hüllermeier.
\newblock {FURIA}: an algorithm for unordered fuzzy rule induction.
\newblock {\em Data Mining and Knowledge Discovery}, 19(3):293--319, Dec. 2009.

\bibitem{junker_quickxplain:_2004}
U.~Junker.
\newblock {QuickXPlain:} {Preferred} explanations and relaxations for
  over-contrained problems.
\newblock In {\em Proceedings of the 19th National Conference on Artifical
  Intelligence}, pages 167--172, 2004.

\bibitem{kampel_survey_2019}
L.~Kampel and D.~E. Simos.
\newblock A survey on the state of the art of complexity problems for covering
  arrays.
\newblock {\em Theoretical Computer Science}, 800:107--124, Dec. 2019.

\bibitem{Kirnap0BECY21}
{\"{O}}.~Kirnap, F.~Diaz, A.~Biega, M.~D. Ekstrand, B.~Carterette, and
  E.~Yilmaz.
\newblock {Estimation of Fair Ranking Metrics with Incomplete Judgments}.
\newblock In {\em Proc.~{WWW}}, 2021.

\bibitem{DBLP:journals/mansci/LambrechtT19}
A.~Lambrecht and C.~Tucker.
\newblock {Algorithmic Bias? An Empirical Study of Apparent Gender-Based
  Discrimination in the Display of {STEM} Career Ads}.
\newblock {\em Management Science}, 65(7):2966--2981, 2019.

\bibitem{lapa_increasing_2024}
K.~Lapa.
\newblock Increasing the explainability and trustiness of {Wang}–{Mendel}
  fuzzy system for classification problems.
\newblock {\em Applied Soft Computing}, 167:112257, Dec. 2024.

\bibitem{lockhart_significance_2014}
R.~Lockhart, J.~Taylor, R.~J. Tibshirani, and R.~Tibshirani.
\newblock A significance test for the lasso.
\newblock {\em The Annals of Statistics}, 42(2):413--468, Apr. 2014.
\newblock Publisher: Institute of Mathematical Statistics.

\bibitem{Mahdavi}
M.~Mahdavi, Z.~Abedjan, R.~Castro~Fernandez, S.~Madden, M.~Ouzzani,
  M.~Stonebraker, and N.~Tang.
\newblock Raha: A configuration-free error detection system.
\newblock In {\em Proceedings of the 2019 International Conference on
  Management of Data}, SIGMOD '19, page 865–882, New York, NY, USA, 2019.
  Association for Computing Machinery.

\bibitem{melchiorre2021investigating}
A.~B. Melchiorre, N.~Rekabsaz, E.~Parada{-}Cabaleiro, S.~Brandl, O.~Lesota, and
  M.~Schedl.
\newblock Investigating gender fairness of recommendation algorithms in the
  music domain.
\newblock {\em Information Processing and Management}, 2021.

\bibitem{mendel_2021_critical}
J.~M. Mendel and P.~P. Bonissone.
\newblock Critical thinking about explainable ai (xai) for rule-based fuzzy
  systems.
\newblock {\em IEEE Transactions on Fuzzy Systems}, 29(12):3579--3593, 2021.

\bibitem{miller2012simultaneous}
R.~Miller.
\newblock {\em Simultaneous Statistical Inference}.
\newblock Springer Series in Statistics. Springer New York, 1981.

\bibitem{pantazis_2018_multipleequivalentsolutionslasso}
Y.~Pantazis, V.~Lagani, P.~Charonyktakis, and I.~Tsamardinos.
\newblock Multiple equivalent solutions for the lasso, 2018.

\bibitem{perera_boston_2018}
P.~Perera.
\newblock {The} {Boston} {Housing} {Dataset}, 2018.
\newblock online:
  \href{https://kaggle.com/code/prasadperera/the-boston-housing-dataset}{https://kaggle.com/code/prasadperera/the-boston-housing-dataset}
  [retrieved: 2024-11-04].

\bibitem{press_numerical_1992}
W.~H. Press, S.~A. Teukolsky, W.~T. Vetterling, and B.~P. Flannery.
\newblock {\em {Numerical} {Recipes} in {C}}.
\newblock Cambridge University Press, second edition, 1992.

\bibitem{RajE22}
A.~Raj and M.~D. Ekstrand.
\newblock Measuring fairness in ranked results: An analytical and empirical
  comparison.
\newblock In E.~Amig{\'{o}}, P.~Castells, J.~Gonzalo, B.~Carterette, J.~S.
  Culpepper, and G.~Kazai, editors, {\em {SIGIR} '22: The 45th International
  {ACM} {SIGIR} Conference on Research and Development in Information
  Retrieval, Madrid, Spain, July 11 - 15, 2022}, pages 726--736. {ACM}, 2022.

\bibitem{reiter_theory_1987}
R.~Reiter.
\newblock {A} theory of diagnosis from first prinicples.
\newblock {\em Artificial Intelligence}, 32(1):57--95, 1987.
\newblock online:
  \href{http://linkinghub.elsevier.com/retrieve/pii/0004370287900622}{http://linkinghub.elsevier.com/retrieve/pii/0004370287900622}
  [retrieved: 2024-11-04].

\bibitem{RekabsazKS21}
N.~Rekabsaz, S.~Kopeinik, and M.~Schedl.
\newblock Societal biases in retrieved contents: Measurement framework and
  adversarial mitigation of {BERT} rankers.
\newblock In {\em Proc.~{ACM SIGIR}}, pages 306--316, 2021.

\bibitem{Rekabsaz0HH21}
N.~Rekabsaz, R.~West, J.~Henderson, and A.~Hanbury.
\newblock Measuring societal biases from text corpora with smoothed first-order
  co-occurrence.
\newblock In {\em Proceedings of the Fifteenth International {AAAI} Conference
  on Web and Social Media, {ICWSM} 2021, held virtually, June 7-10, 2021},
  pages 549--560. {AAAI} Press, 2021.

\bibitem{ross_fuzzy_2002}
T.~Ross, J.~M. Booker, and W.~J. Parkinson.
\newblock {\em {Fuzzy} {Logic} and {Probability} {Applications:} {Bridging} the
  gap}.
\newblock ASA SIAM, 2002.

\bibitem{ross_1994_FuzzyLW}
T.~J. Ross.
\newblock {\em Fuzzy Logic with Engineering Applications}.
\newblock Wiley, 1994.

\bibitem{tibshirani_1996_RegressionSA}
R.~Tibshirani.
\newblock Regression shrinkage and selection via the lasso.
\newblock {\em Journal of the Royal Statistical Society. Series B
  (Methodological)}, 58(1):267--288, 1996.

\bibitem{wang_generating_1992}
L.-X. Wang and J.~Mendel.
\newblock Generating fuzzy rules by learning from examples.
\newblock {\em IEEE Transactions on Systems, Man, and Cybernetics},
  22(6):1414--1427, Dec. 1992.

\bibitem{wang_factors_2020}
R.~Wang, F.~M. Harper, and H.~Zhu.
\newblock Factors influencing perceived fairness in algorithmic
  decision-making: Algorithm outcomes, development procedures, and individual
  differences.
\newblock In {\em Proceedings of the 2020 CHI Conference on Human Factors in
  Computing Systems}, CHI '20, page 1–14, New York, NY, USA, 2020.
  Association for Computing Machinery.

\bibitem{YaoH17}
S.~Yao and B.~Huang.
\newblock Beyond parity: Fairness objectives for collaborative filtering.
\newblock In {\em Proc. of NeurIPS}, pages 2921--2930, 2017.

\bibitem{zhang_mitigating_2018}
B.~H. Zhang, B.~Lemoine, and M.~Mitchell.
\newblock Mitigating {Unwanted} {Biases} with {Adversarial} {Learning}.
\newblock In {\em Proceedings of the 2018 {AAAI}/{ACM} {Conference} on {AI},
  {Ethics}, and {Society}}, pages 335--340, New Orleans LA USA, Dec. 2018. ACM.

\end{thebibliography}

%
%
%
%
%
%
%
%

\end{document}